\documentclass[acmtog,timestamp,final]{acmart}

\usepackage{booktabs} 
\usepackage[ruled]{algorithm2e} 
\usepackage{color}

\SetAlFnt{\small}
\SetAlCapFnt{\small}
\SetAlCapNameFnt{\small}
\SetAlCapHSkip{0pt}
\IncMargin{-\parindent}

\acmJournal{TOG}
\acmYear{2018}
\acmVolume{37}
\acmNumber{6}
\acmArticle{260}
\acmMonth{11}

\setcopyright{none}

\acmDOI{10.1145/3272127.3275060}
\acmISBN{0730-0301}

\setcitestyle{square}

\begin{document}
\title{Image Super-Resolution via Deterministic-Stochastic Synthesis and Local Statistical Rectification}
\author{Weifeng Ge}
\authornote{Both authors contributed equally to this paper.}
\author{Bingchen Gong}
\authornotemark[1]
\author{Yizhou Yu}
\affiliation{%
  \institution{The University of Hong Kong}
  \department{Department of Computer Science}
  \city{Hong Kong}
}

\begin{teaserfigure}
\setlength{\abovecaptionskip}{5pt}
\setlength{\belowcaptionskip}{9pt}
  \centering
  \includegraphics[width=1.0\linewidth]{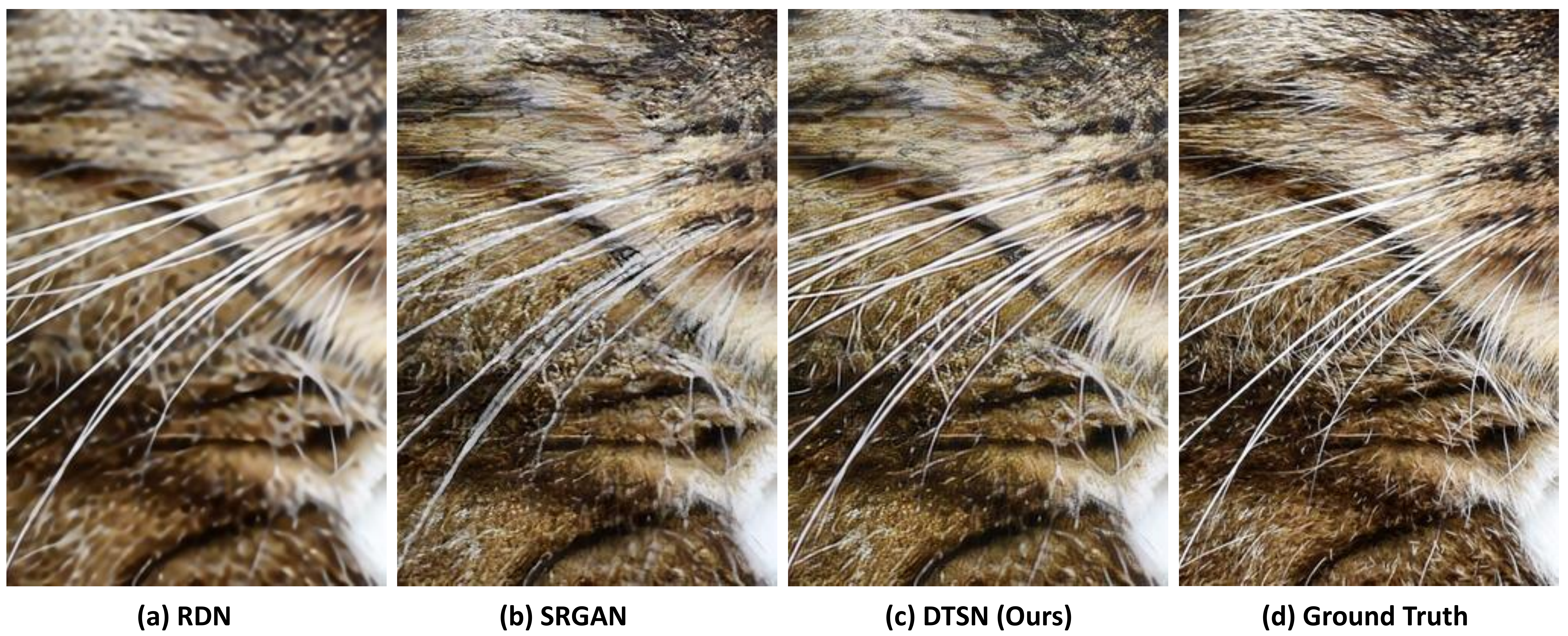}
  \caption{Comparison of visual results from state-of-the-art algorithms for $\times 4$ single image superresolution. (a) RDN~\cite{zhang2018residual}: State-of-the-art algorithm under PSNR/SSIM. (b) SRGAN~\cite{Ledig_2017_CVPR}: Sate-of-the-art algorithm in visual quality. (c) DTSN (Deterministic-sTochastic Synthesis Net): the proposed method. (d) High-resolution ground truth.}
  \label{fig:teaser}
\end{teaserfigure}

\begin{abstract}
Single image superresolution has been a popular research topic in the last two decades and has recently received a new wave of interest due to deep neural networks. In this paper, we approach this problem from a different perspective. With respect to a downsampled low resolution image, we model a high resolution image as a combination of two components, a deterministic component and a stochastic component. The deterministic component can be recovered from the low-frequency signals in the downsampled image. The stochastic component, on the other hand, contains the signals that have little correlation with the low resolution image. We adopt two complementary methods for generating these two components. While generative adversarial networks are used for the stochastic component, deterministic component reconstruction is formulated as a regression problem solved using deep neural networks. Since the deterministic component exhibits clearer local orientations, we design novel loss functions tailored for such properties for training the deep regression network. These two methods are first applied to the entire input image to produce two distinct high-resolution images. Afterwards, these two images are fused together using another deep neural network that also performs local statistical rectification, which tries to make the local statistics of the fused image match the same local statistics of the groundtruth image. Quantitative results and a user study indicate that the proposed method outperforms existing state-of-the-art algorithms with a clear margin.
\end{abstract}

%
%

%

\maketitle

\begin{figure}[ht]
  \centering
  \includegraphics[width=1.0\linewidth]{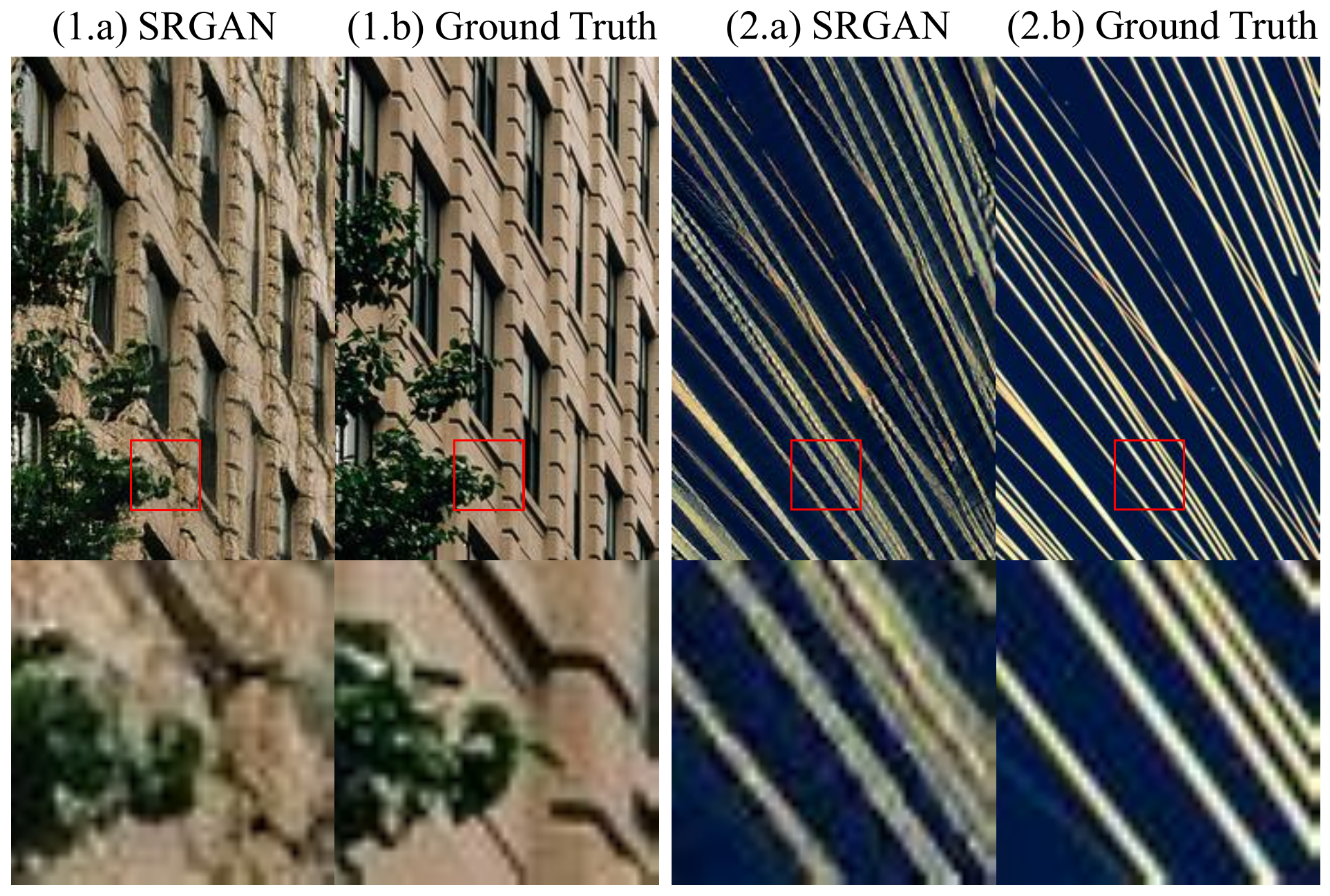}
  \caption{Examples of the stochastic components generated by SRGAN~\cite{Ledig_2017_CVPR}. SRGAN may destroy structures while producing too many stochastic details in certain areas.}
  \label{Fig:chaos in image structure}
\end{figure}

\section{Introduction}
Single image super-resolution (SISR) aims to estimate high resolution images from low resolution inputs. It has been a popular research topic in the last two decades and has recently received a new wave of interest due to deep neural networks. Image super-resolution has been used for many different applications, including facial image improvement~\cite{BK2000,HLQS2010}, compressed image/video enhancement~\cite{SMK2003,GAM2004}, high dynamic range imaging~\cite{BGVL2012}, image mosaicing~\cite{CZ1998} as well as satellite and aerial imaging~\cite{YG2012}.

It is well known that single image super-resolution is ill-posed because certain high frequency signals in the high resolution image is irrecoverably lost in the corresponding low resolution image due to downsampling. Thus SISR does not have a unique solution, and many slightly different high resolution images could become the same low resolution image after downsampling. Therefore, a high resolution image can be decomposed into two components, a {\em deterministic component} and a {\em stochastic component}. The deterministic component contains all the low-frequency signals as well as part of the high-frequency signals highly correlated to the low-frequency signals, which implies that the high-frequency signals in the deterministic component could still be recovered from the low-frequency signals in a downsampled image. The stochastic component, on the other hand, contains the rest of the signals that have little correlation with the low-frequency signals. It is the stochastic component that makes image super-resolution ill-posed.

According to the above analysis, the biggest challenges in natural image super-resolution is two-fold: first, how to reconstruct the deterministic components in the high resolution image from the low resolution image? second, how to hallucinate plausible stochastic components for the high resolution image which is compatible with the low resolution image?

To tackle these challenges, it is necessary to distinguish salient curvilinear structures, such as edges and contours, from the remaining pixels. Curvilinear structures exhibit clear and spatially coherent orientations while the orientation at other pixels, including those in texture regions, is more ambiguous or spatially incoherent. This distinction implies that the deterministic component plays a much more dominant role than the stochastic component around curvilinear structures while the stochastic component becomes more important at other pixels. We call pixels on curvilinear structures {\em structural pixels}, and the rest of the pixels {\em non-structural pixels}.

In this paper, we approach single image super-resolution from a different perspective and propose novel solutions to the aforementioned challenges. Our solution contains three cascaded stages. In the first stage, we perform deterministic component reconstruction and stochastic component hallucination independently.
Deterministic component reconstruction is formulated as a regression problem solved using deep neural networks because high-frequency signals in the deterministic component are strongly correlated to signals in the low-resolution image.
We design novel loss functions for training the deep regression network.
Specifically, in addition to the mean absolute error (MAE) typically used for regression networks, we introduce a generalized gradient loss for better preserving local variations as well as an orientation loss that reconstructs local orientations.
The orientation loss is particularly useful for structural pixels because the deterministic component at such pixels exhibits clear local orientations.
For stochastic component hallucination, generative adversarial networks (GANs~\cite{goodfellow2014generative}) are adopted to generate natural high frequency signals.
Because GANs~\cite{Ledig_2017_CVPR} usually introduce artifacts that destroy image structures, they are better suited for non-structural pixels.

Although the above two methods are complementary and better suited for different subsets of pixels, both of them are applied to the entire input image to produce a pair of deterministic and stochastic components for every pixel. In the second stage, these two high resolution components are fused together to produce a single image, which should mostly preserve the deterministic component only at structural pixels and combine the two components at non-structural pixels. Since structural pixels are unknown at test time, we train another deep neural network to implicitly label structural pixels and complete the fusion. The input to this network consists of the original low-resolution input image and the aforementioned two high resolution images generated from the two complementary methods, respectively.

Although GANs can be used for generating natural stochastic components, the synthesis process is a global operation and is incapable of generating stochastic components with spatially varying statistical properties that accurately match the local deterministic component.
Therefore, in the last stage, we rectify deviations introduced into the stochastic component in earlier stages by matching local statistics of the fused image with the same local statistics of the groundtruth high resolution image. Even though matching local statistics does not enforce uniqueness at the pixel level, it serves as an as-strong-as-possible constraint due to the ill-posedness of the problem. Inspired by deep learning based style transfer~\cite{gatys2016image} and texture synthesis~\cite{sendik2017deep}, local Gram matrices and local correlation matrices are adopted as local statistics.

We have evaluated our method on a few popular benchmark datasets. Quantitative experimental results demonstrate that our deterministic component reconstruction algorithm achieves state-of-the-art PSNR and SSIM on Set5~\cite{bevilacqua2012low}, Set14~\cite{zeyde2010single}, B100~\cite{martin2001database}, Urban100~\cite{huang2015single} and DIV2K validation set~\cite{Timofte_2017_CVPR_Workshops}. On the other hand, the synthesized stochastic component may be either perceptually or statistically similar to the stochastic details in the ground truth, but does not have strict pixelwise correspondences with the latter. Therefore, subjective evaluation based on user studies is better suited than quantitative measures based on pixel-level registration for evaluating image superresolution results with stochastic components. To this end, a user study with 42 participants concludes that in comparison to existing state-of-the-art superresolution algorithms, our proposed method is capable of generating high resolution images with significantly better visual quality.

In summary, this paper has the following contributions:
{\flushleft $\bullet$}
A novel pipeline for single image super-resolution is introduced by decomposing SISR into three consecutive stages, deterministic component reconstruction and stochastic component hallucination, deterministic and stochastic component fusion, and local statistical rectification.

{\flushleft $\bullet$} A deep learning algorithm is designed for deterministic component reconstruction. Novel loss functions, including a generalized gradient loss and an orientation loss, are proposed to enforce properties of the deterministic component.

{\flushleft $\bullet$} A deep learning algorithm is designed for image fusion as well as local statistical rectification.
Novel loss functions based on local Gramm matrices and local correlation matrices are proposed to make local statistics of the generated high resolution image match the same local statistics of the groundtruth high resolution image.

\section{Related Work}
\noindent\textbf{Exemplar or Dictionary Based Super-Resolution.} Exemplar-based methods require a database of external images, and synthesize a high-resolution version of the input image by searching for exemplars in the image database and transferring relevant patches from the retrieved exemplars~\cite{freeman2002example,chang2004super}. To synthesize high-quality results, the image database needs to be fairly large, increasing the computational cost for exemplar and patch-level retrieval. Dictionary-based methods aim to mitigate this problem by learning a compact dictionary in a feature space~\cite{timofte2013anchored,timofte2014a+}. Alternatively, over-complete dictionaries are learned in methods based on the sparse signal reconstruction theory~\cite{jianchao2008image,zeyde2010single}.
Such methods build over-complete dictionaries for both low-resolution and high-resolution images, and make certain assumptions about the sparse coefficients for the dictionaries in both domains.
Because there could be multiple high resolution images corresponding to a given low resolution image, dictionary-based methods in essence average multiple possible solutions and usually generate blurry results.

Almost all exemplar-based super-resolution algorithms need a high-resolution initial image upsampling the low-resolution input image. Edge-based methods~\cite{fattal2007, zhou2011} can be used to produce better initial images. Tai {\em et al.}~\shortcite{tai2010super} further combined edge-based methods with exemplar-based synthesis. There exist major differences between our proposed method and the algorithm in \cite{tai2010super}. First, the deterministic component in our model exists everywhere, including texture regions, and not just at edge pixels. Second, while Tai {\em et al.}~\shortcite{tai2010super} treat all edge pixels equally, edge pixels in texture regions are not structural pixels in our method because they do not exhibit spatial coherence to form lines or curves.  In addition to such conceptual differences, at the algorithmic level, our method is based on deep neural networks and learning while Tai {\em et al.}~\shortcite{tai2010super} rely on exemplar-based synthesis.

\noindent\textbf{Deep Learning Based Super-Resolution.} Given pairs of low-resolution and high-resolution images, super-resolution can be naturally cast as a regression problem that maps low-resolution images to their corresponding high-resolution images. On the other hand, deep neural networks are well known for their strong regression capability. Therefore, in recent years, many researchers have approached image super-resolution using deep learning. SRCNN~\cite{dong2014learning} is a deep network with three convolutional layers designed for super-resolution regression. Other researchers have focused on designing better network architectures and more powerful loss functions. Unlike SRCNN, VDSR~\cite{kim2016accurate} learns the difference between the high-resolution image and the low-resolution image with a much deeper convolutional network. SRResNet~\cite{Ledig_2017_CVPR} shows that multiple concatenated residual blocks can further improve the regression performance. EDSR~\cite{lim2017enhanced} uses very deep and wide residual networks and the mean absolute error for both the original images and the image gradients. LapSRN~\cite{LapSRN} uses both residual networks and the Laplacian pyramid to perform regression at multiple intermediate resolutions. Although using the mean absolute error or the mean squared error as the training loss favors numerical measures, such as peak signal-to-noise ratio (PSNR), such loss functions do not emphasize the visual quality of the results.

As discussed earlier, SISR is ill-posed and only the high-resolution deterministic component can be uniquely reconstructed from the low-resolution image. Although the resulting images have high-resolution structures, they lack interesting high-frequency details from the stochastic component. Therefore, regression alone is unable to produce results that closely resemble natural images. To this end, generative adversarial networks (GANs) can help to a certain extent. SRGAN~\cite{Ledig_2017_CVPR} incorporates a perception-friendly discriminator that is trained to tell what kind of high-frequency details look natural. High-resolution images generated from SRGAN do have higher visual quality than those from regression only networks. Nevertheless, GANs do not completely solve the problem. They can only broadly tell whether certain high-frequency details look natural, but cannot tell whether such details are consistent with the given low-resolution input, or more specifically, consistent with every local region in the low-resolution input. This is the reason why SRGAN often generates high-resolution images where certain high-frequency details do not appear to be in the right place. In addition, GAN sometimes generates naturally looking high-frequency details at the cost of destroying existing image structures in the low-resolution input.

In contrast, the method proposed in this paper handles structural pixels and non-structural pixels in a more balanced way. To preserve image structures, high-resolution structural pixels are still reconstructed using regression but with additional novel loss functions. The synthesized stochastic component not only looks natural, but also is consistent with local contents in the low-resolution input.


\noindent\textbf{Texture Enhancement Based Super Resolution.} Previous work exists on generating high-resolution textures for SISR. Huang {\em et al.}~\shortcite{huang2015single} exploit self-similarity of an image at different scales and adopt a patch-based transformation model to handle geometric transformations. Ahn {\em et al.}~\shortcite{ahn2016texture} first generate a high-resolution texture and then use this texture image to produce another high-resolution style image with scaling and tiling. Finally they apply a style transfer algorithm to transfer the high-resolution style to the target image. Sun {\em et al.}~\shortcite{sun2017super} complete main image structures with user interactions and then generate textures via structure propagation. They treat SISR as a constrained texture transfer problem, and need an example texture/style image when hallucinating high-frequency image details.

High-frequency details in natural images are usually complex and spatially varying. Therefore, existing SISR methods based on texture enhancement are not well suited for the intended task because they either use homogeneous textures or fail to specify the spatial correspondences between the texture style image and the estimated high-resolution image. In our proposed method, we aim to synthesize high-resolution stochastic details that satisfy spatially varying statistical properties defined by local Gram matrices and local correlation matrices.


\begin{figure*}[ht]
  \centering
  \includegraphics[width=1.0\linewidth]{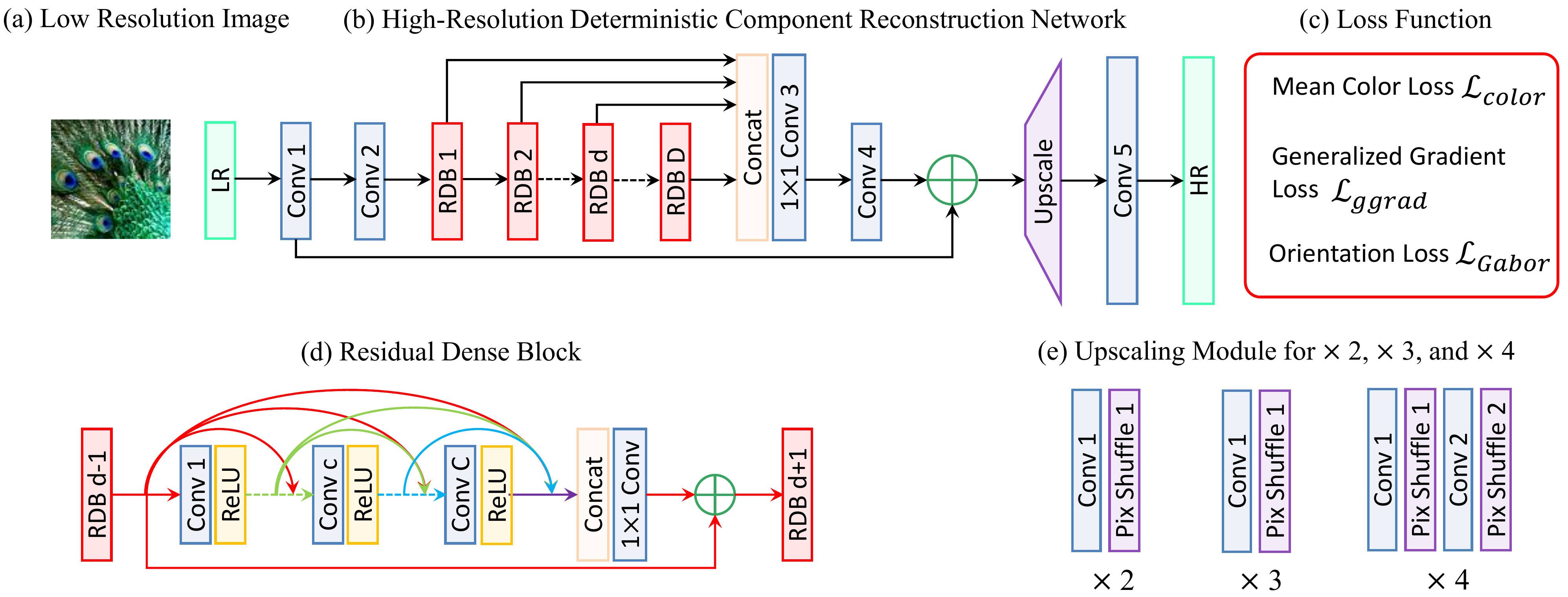}
  \caption{Pipeline for high resolution deterministic component reconstruction. (a) Low resolution input image $\boldsymbol{\mathcal{I}}^{lr}$. (b) The deterministic component reconstruction network $\phi_{d}\left ( \cdot,{\theta}_d \right )$, which is composed of 4 convolutional layers, 1 concatenation layer, 1 upsampling module, and 16 residual dense blocks. (c) Training loss $\mathcal{L}_{d}$, which contains a mean color loss $\mathcal{L}_{color}$, a generalized gradient loss $\mathcal{L}_{ggrad}$ and an orientation loss $\mathcal{L}_{Gabor}$. (d) A residual dense block. (e) The upscaling modules for $\times$ 2, $\times$ 3 and $\times$ 4 superresolution.}
  \label{Fig:image structure}
\end{figure*}

\noindent\textbf{Style Transfer and Texture Synthesis.} Gatys {\em et al.}~\shortcite{gatys2016image} use global Gram matrices computed for a subset of feature maps in a convolutional neural network to represent the style of an image. They transfer the style from a style image to a target image by minimizing a content loss and a style loss. There has been much work that adopts the same style representation while improving various other aspects, such as efficiency. Sendik {\em et al.}~\shortcite{sendik2017deep} use a global correlation matrix computed for the top convolutional layer in a deep neural network to perform texture synthesis and achieve state-of-the-art results. In contrast, to be able to rectify high-frequency details with spatially varying statistics, we generalize the global Gram matrix and global correlation matrix used in such work to densely computed local Gram matrices and local correlation matrices during local statistical rectification of high-resolution stochastic components.

\vspace{-2mm}

\section{Overview}
Given a low-resolution image as input, SISR aims to estimate a high-resolution counterpart that meets the following two requirements: 1) it looks like a natural image, and 2) its downsampled version is the same as the low-resolution input. For SISR based on supervised learning, one training image pair contains a high-resolution image $\boldsymbol{\mathcal{I}}^{hr}$ and its downsampled low-resolution version (usually bicubic interpolation) $\boldsymbol{\mathcal{I}}^{lr}$. In our approach, SISR is decomposed into three sequential stages: the first stage performs high-resolution reconstruction of the deterministic component $\boldsymbol{\mathcal{I}}^{hr,d}$ and high-resolution hallucination of the stochastic component $\boldsymbol{\mathcal{I}}^{hr,t}$ independently; the second stage fuses the high-resolution deterministic and stochastic components generated in the previous stage; in the final stage, spatially varying local statistics are introduced to rectify the the stochastic component in the fused image, and produces the final high-resolution image.

\section{High-Resolution Deterministic Component Reconstruction}

\subsection{Network Architecture} As shown in Fig.~\ref{Fig:image structure}, we adopt an existing deep network, called residual dense network (RDN~\cite{zhang2018residual}) for deterministic component reconstruction.
This network is composed of 4 convolutional layers, 1 upscaling module, and 16 residual dense blocks (RDBs).
The detailed structure inside each module is also shown in Fig. \ref{Fig:image structure}.
Each residual dense block has local dense connections among all of its convolutional layers. It takes the features from the preceding RDB as the input, and feed these features to all of its own convolutional layers. Each convolutional layer in a residual dense block also receives features from all the preceding convolutional layers in the same RDB.
In addition, RDBs adopt the global residual learning mechanism in SRResNet~\cite{Ledig_2017_CVPR} to combine shallow features near the input and deep features near the output together, resulting in better regression ability.

\subsection{Training Loss}
In addition to the conventional color loss, we design novel loss functions for training the above deep network.
Specifically, we introduce a generalized gradient loss for better preserving local variations and an orientation loss that reconstructs local orientations. The generalized gradient loss suppresses noise while preserving gradients. The orientation loss preserves curvilinear structures, including lines and curves.

\paragraph{Color Loss} To make the reconstructed image match the pixel colors of the groundtruth high-resolution image $\boldsymbol{\mathcal{I}}^{hr}$, we use their mean absolute difference ($\mathcal{L}_1$ norm) as the color loss, which is defined as follows.
    \begin{equation}
       \mathcal{L}_{color} =   \frac{1}{Z_c} \sum_{m,n} \left\| \boldsymbol{\mathcal{I}}^{hr,d}_{m,n} - \boldsymbol{\mathcal{I}}^{hr}_{m,n} \right\|_1,
    \label{eq:image content loss}
    \end{equation}
where $\boldsymbol{\mathcal{I}}^{hr,d}_{m,n}$ is the color value vector at ($m,n$) in $\boldsymbol{\mathcal{I}}^{hr,d}$, $\boldsymbol{\mathcal{I}}^{hr}_{m,n}$ is the color value vector at ($m,n$) in $\boldsymbol{\mathcal{I}}^{hr}$,  $Z_c = 3MN$, $M$ and $N$ are the numbers of rows and columns of $\boldsymbol{I}^{hr}$, and $3$ is the number of color channels.

\paragraph{Generalized Gradient Loss} To emphasize edges and local structures during image reconstruction, we design a generalized gradient loss. Inspired by handcrafted features, such as local binary patterns (LBP), which compare pairs of nearby pixels, a generalized gradient vector at pixel ($m$, $n$) is defined as a feature vector holding pairwise differences between $\boldsymbol{\mathcal{I}}$($m, n$) and all other pixels in its $r \times r$ neighborhood. This feature is called generalized gradient because a conventional gradient for a discrete image is defined as a vector with two pairwise differences. The generalized gradient includes more structural information because it can use a larger neighborhood and a much larger number of pixel pairs. To improve the quality of deterministic component reconstruction, we wish the generalized gradients in $\boldsymbol{\mathcal{I}}^{hr,d}$ to be as close to the corresponding generalized gradients in $\boldsymbol{\mathcal{I}}^{hr}$ as possible.

Thus, the generalized gradient loss $\mathcal{L}_{ggrad}$ is defined as follows.
    \begin{equation}
      \mathcal{L}_{ggrad} =   \frac{1}{Z_g} \sum _{m,n} \sum _{i,j \in R_{mn}} \left\| \left(\boldsymbol{\mathcal{I}}^{hr,d}_{m,n} - \boldsymbol{\mathcal{I}}^{hr,d}_{i,j}\right) - \left(\boldsymbol{\mathcal{I}}^{hr}_{m,n} - \boldsymbol{\mathcal{I}}^{hr}_{i,j}\right) \right\|_1,
    \label{eq:edge preserving loss}
    \end{equation}
where $Z_g = 3MN \left( r^2 - 1 \right)$, and $R_{mn}$ is a neighborhood of the pixel at ($m$, $n$). Neighborhood size is set to $15 \times 15$ ($r=15$).


\paragraph{Orientation Loss} Gabor filters are widely used in image processing because of its strong ability in texture analysis. In the spatial domain, a 2D Gabor filter is essentially a Gaussian kernel modulated by a sinusoidal plane wave. By modifying the standard deviation of the Gaussian kernels in the two principal directions or changing the frequency and phase of the sinusoidal plane wave, we can obtain a set of 2D Gabor filters. Here we follow the parameter setting used in \cite{ahn2016texture}. Our Gabor filters have 4 scales and a dense set of 18 orientations. We set up a filter every 10 degrees to accurately characterize curvilinear structures with different orientations. In addition, we adopt three different ratios between the standard deviations in the two principal directions--$1:1$, $1:2$, and $2:1$. When both real and imaginary parts are used, in total, there are 432 Gabor kernels. Supplementary materials show a subset of these filters.

The orientation loss $\mathcal{L}_{Gabor}$ measures the mean difference between corresponding Gabor filter responses once they have been applied to both $\boldsymbol{\mathcal{I}}^{hr,d}$ and $\boldsymbol{\mathcal{I}}^{hr}$, and can be written as follows.
    \begin{equation}
       \mathcal{L}_{Gabor} =   \frac{1}{Z_G} \sum _{m,n} \left\| G(\boldsymbol{\mathcal{I}}^{hr,d})_{m,n} - G(\boldsymbol{\mathcal{I}}^{hr})_{m,n} \right\|_1,
    \label{eq:Gabor oriented loss}
    \end{equation}
where $C_G=432$ is the number of Gabor filters, $Z_G = 3C_G MN$, and $G(\cdot)$ represents the concatenation of Gabor filter response maps of $3$ image channels. In our experiments, the spatial support of a Gabor filter is a local image patch with $51$x$51$ pixels.

\paragraph{Final Training Loss} The final training loss for deterministic component reconstruction is a weighted sum of all three previously defined losses.
    \begin{equation}
       \mathcal{L}_d = \mathcal{L}_{color} + \alpha_1 \mathcal{L}_{ggrad} + \alpha_2 \mathcal{L}_{Gabor},
    \label{eq:Gabor oriented loss}
    \end{equation}
where $\alpha_1 = 1$ and $\alpha_2 = 1$ in all our experiments. These parameters are set to make the gradients of the three losses have similar magnitude.

\begin{figure*}[ht]
  \centering
  \includegraphics[width=1.0\linewidth]{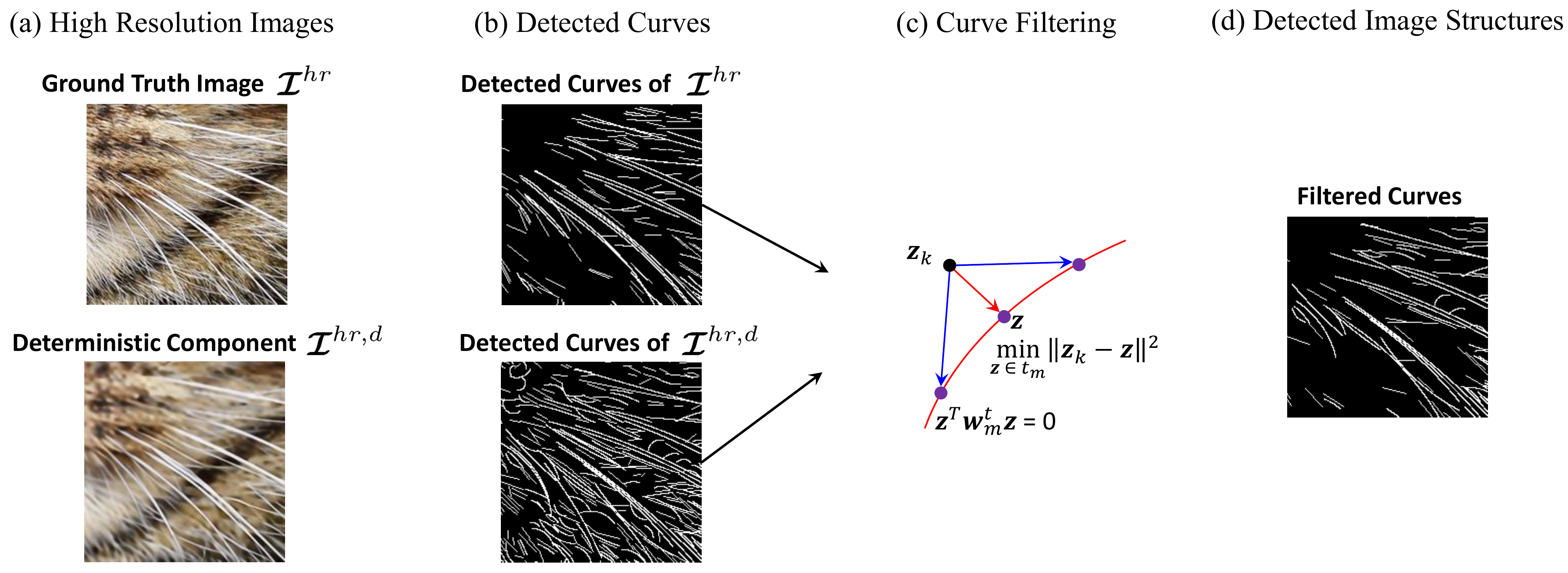}
  \caption{Pipeline for image structure detection. Form left to right. (a) The ground truth image $\boldsymbol{\mathcal{I}}^{hr}$ and the high resolution deterministic component $\boldsymbol{\mathcal{I}}^{hr,d}$. (b) Detected curves in GT image and the high resolution deterministic component. (c) Curve filtering. (d) Image structure detection by curve filtering.}
  \label{Fig:strcture detection}
\end{figure*}

\section{High-Resolution Stochastic Component Hallucination}
Generative adversarial networks~\cite{goodfellow2014generative} have been successfully incorporated into recent image superresolution algorithms~\cite{Ledig_2017_CVPR}, which are capable of generating natural high-resolution details by exploiting a combination of color loss, perceptual loss and adversarial loss in the training process. However, we have observed that the adversarial loss cannot well preserve curvilinear structures with spatially coherent orientations, as shown in Fig~\ref{Fig:chaos in image structure}. This is due to the fact that the architecture of GANs is not tailored for modeling spatially varying long-range correlations. Because of this, we limit the application of GANs to high resolution stochastic component hallucination only. In practice, we use the SRGAN algorithm in \cite{Ledig_2017_CVPR}.
Note that in the loss functions for SRGAN, the color loss is for reconstructing the deterministic component, and the perceptual loss and adversarial loss are used for generating the stochastic component. As a result, results from SRGAN have both deterministic and stochastic components. To extract the stochastic component, we obtain the deterministic component from a second network trained with the color loss only, and subtract the deterministic component from the final image generated from SRGAN. The second network shares the same architecture as the generator network in SRGAN.



\section{Deterministic and Stochastic Component Fusion}
We need to fuse the two complementary components generated in the previous stage to generate a single high-resolution image.
In general, every pixel in the fused image should be a weighted sum of these two components.
Since the deterministic component can be used for all pixels while the stochastic component from SRGAN is suited for non-structural pixels only, the fused image should take the deterministic component only at structural pixels and combine the two components at non-structural pixels. Since groundtruth structural pixels are unknown at test time, we cannot directly generate the fused image by following this observation. Fortunately, structural pixels can be explicitly detected in groundtruth high resolution images. A mask of structural pixels for every groundtruth image can be incorporated into the training data for a deep neural network, which implicitly labels structural pixels and completes the fusion at test time.

\subsection{Network Architecture}
\begin{figure*}[ht]
  \centering
  \includegraphics[width=1.0\linewidth]{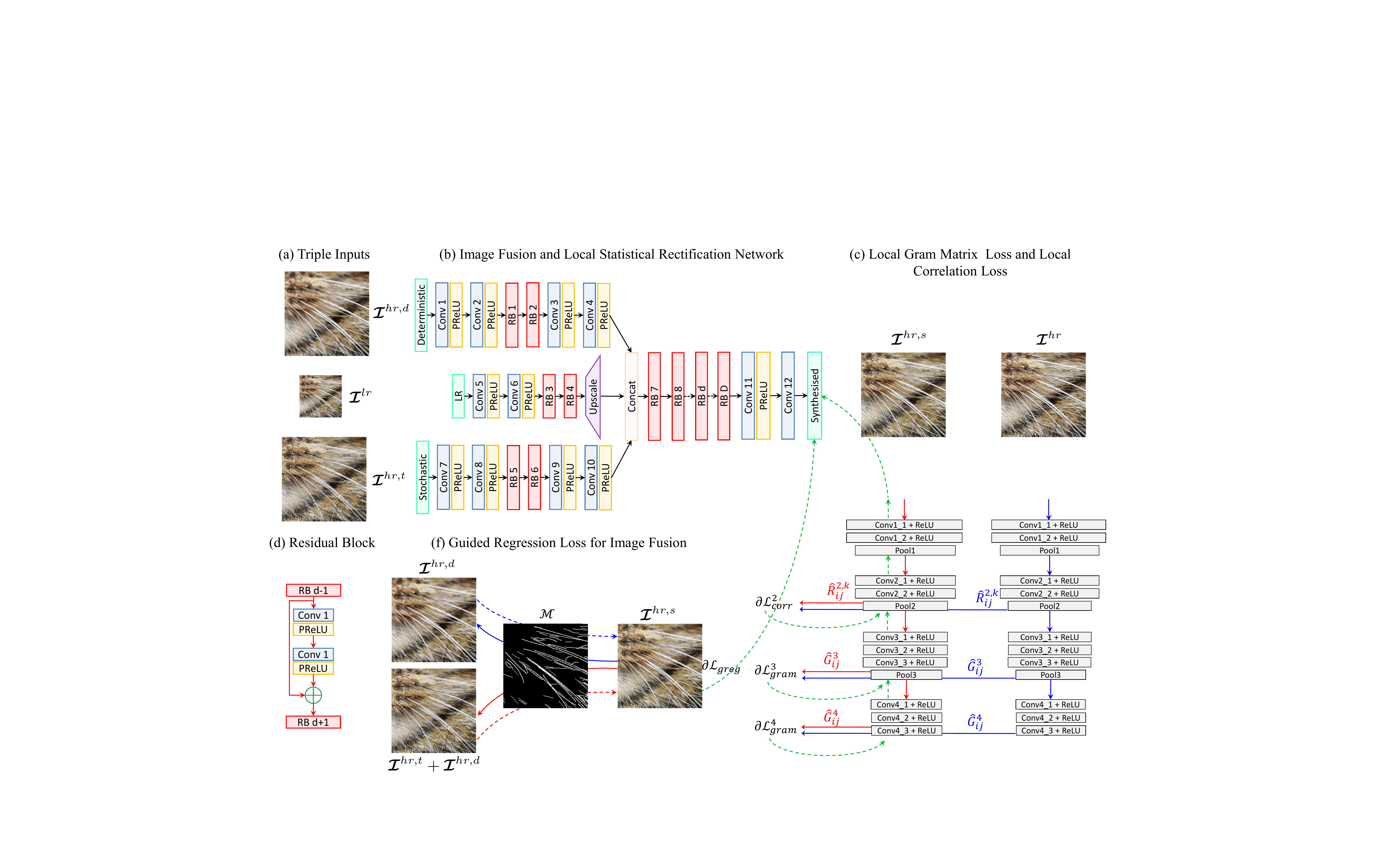}
  \caption{Pipeline for image fusion and local statistical rectification. (a) Deterministic component, stochastic component and low resolution input image. (b) Image fusion and local statistical rectification network. (c) The local Gram matrix loss $\mathcal{L}_{Gram}$, and the local correlation matrix loss $\mathcal{L}_{corr}$ in the combined loss $\mathcal{L}_{t}$. (d) A residual block. (e) The guided regression loss $\mathcal{L}_{greg}$ in the combined loss $\mathcal{L}_{t}$.}
  \label{Fig:synthesis network}
\end{figure*}

As shown in Figure~\ref{Fig:synthesis network}, our deep network for deterministic and stochastic component fusion, $\phi_{s}\left ( \cdot,{\theta}_s \right )$, is composed of convolution layers, residual blocks and upscaling layers. Each residual block has two convolutional layers. There are 16 residual blocks in total and the nonlinear activation function is PReLU~\cite{he2015delving}. The input to this deep network consists of three parts: the original low-resolution input image $\mathcal{I}^{lr}$  and two high resolution images. The first high resolution image, $\boldsymbol{\mathcal{I}}^{hr,d}$, is generated from the deterministic component reconstruction network, and the other high resolution image, $\boldsymbol{\mathcal{I}}^{hr,t} + \boldsymbol{\mathcal{I}}^{hr,d}$, is the stochastic component generated using SRGAN~\cite{Ledig_2017_CVPR}.



\subsection{Network Training} To make the above deep network produce correct image fusion results, we need to prepare a training set. Each training sample in the training set includes the aforementioned three input images and a corresponding binary mask separating structural pixels from non-structural ones. This binary mask is only used in the loss function for training the network. The loss function evaluates the quality of the fused image produced by the network. Pixel colors at structural pixels of the fused image should match those from the first high resolution image fed to the network (the deterministic component) while pixel colors at non-structural pixels of the fused image should be the sum of those from the two high resolution images (deterministic + stochastic components). The binary mask includes all pixels in the $5 \times 5$ neighborhood of any image structures detected in the next section.
Given such a training set, the guided regression loss, $\mathcal{L}_{greg}$, for training the fusion network can be defined as follows.
    \begin{equation}
    \begin{aligned}
        \mathcal{L}_{greg} = & \sum _ {m,n} \frac{\mathcal{M}_{m,n}}{3 m_0} \left \| \boldsymbol{\mathcal{I}}^{hr,s}_{m,n} - \left( \boldsymbol{\mathcal{I}}^{hr,d}_{m,n} + \boldsymbol{\mathcal{I}}^{hr,t}_{m,n} \right) \right \| _1 + \\
                   & \sum _ {m,n} \frac{1 - \mathcal{M}_{m,n}}{3 m_1} \left \| \boldsymbol{\mathcal{I}}^{hr,s}_{m,n} - \boldsymbol{\mathcal{I}}^{hr,d}_{m,n} \right \|_1\\
    \end{aligned}
    \end{equation}
where $\boldsymbol{\mathcal{I}}^{hr,s}$ represents the fused image, $\mathcal{M}$ is the binary mask ($\mathcal{M}(x,y) = 0$ means the pixel at $(x,y)$ is a structural pixel),
$m_c$ denotes the number of color channels, $m_0$ is the total number of structural pixels, and $m_1$ is the total number of non-structural pixels.

\subsection{Image Structure Detection} Image structure detection aims to identify important curvilinear structures that define the content and layout of an image. Most of these structures lie along object boundaries. However, traditional edge detection produces too many unimportant intensity/color changes, especially in the presence of high frequency textures. Deep learning based methods, such as HED~\cite{xie2015holistically} and COB~\cite{maninis2018convolutional}, actually perform edge detection according to changes in local semantic meaning. They cannot accurately locate edges and contours, thus cannot be used in image structure detection. ELSD~\cite{puatruaucean2012parameterless} detects and vectorizes curvilinear image structures, including line segments, circular arcs and elliptic arcs, by fitting relevant conic sections to detected edges. In this paper, we identify structural pixels through the primitives vectorized by ELSD.


Since the deterministic component is reconstructed from a low resolution image, not all curvilinear structures in the groundtruth high resolution image belong to the deterministic component. Structural pixels are defined as pixels on the groundtruth curvilinear structures that can be reconstructed from the low resolution image. This definition implies a structural pixel should lie on vectorized curves in two high resolution images, the groundtruth high resolution image and the high resolution image generated from the deterministic component reconstruction network. Thus, structural pixel labeling can be achieved as follows. First, use ELSD to detect and vectorize curves in both aforementioned high resolution images. Second, identify pixels lying on both sets of vectorized curves as structural pixels.

Let $\mathcal{S} = \left \{ s_{i} \right \}^{N_{\mathcal{S}}}_{i=1}$ be the set of vectorized curves in the high resolution image, $\boldsymbol{\mathcal{I}}^{hr,d}$, from the deterministic component reconstruction network. For a curve $s_{i}$ in $\mathcal{S}$, its analytic expression can be written as follows.
    \begin{equation}
[x ~ y ~ 1]\begin{bmatrix}
 w^{11}_i & w^{12}_i & w^{13}_i \\
 w^{21}_i & w^{22}_i & w^{23}_i \\
 w^{31}_i & w^{32}_i & w^{33}_i
\end{bmatrix}\begin{bmatrix}
x\\
y\\
1
\end{bmatrix} = 0,
    \label{eq:perceptual loss}
    \end{equation}
where $x_1 \leq  x \leq  x_2$, $y_1 \leq  y \leq  y_2$. $\left ( x_1,~y_1 \right )$ is the starting point, and  $\left ( x_2,~y_2 \right )$ is the end point. $\textbf{w}_{i}$ is the $3 \times 3$ parameter matrix defining the $i$-th curve $s_{i}$ in $\mathcal{S}$.

Let $\mathcal{T} = \left \{ t_{i} \right \}^{N_{\mathcal{T}}}_{i}$ be the set of vectorized curves in the groundtruth high resolution image. Since curves in $\mathcal{S}$ may not necessarily appear in $\mathcal{T}$, to obtain the subset of image structures that can be reconstructed from low resolution images, we identify the subset of curves in $\mathcal{S}$ that lie closely to curves in $\mathcal{T}$. The closeness score $\sigma_{i}$ of a curve $s_{i}$ in $\mathcal{S}$, when it is compared with curves in $\mathcal{T}$, is calculated as follows.
    \begin{equation}
       f(\textbf{z}_k; \textbf{w}_m^t) = \min _ {\textbf{z} \in t_m} \left \| \textbf{z}_k - \textbf{z} \right \| ^ 2 ,
    \label{eq:perceptual loss}
    \end{equation}
    \begin{equation}
      \sigma _{i} = \frac{1}{K_{i}} \sum _ {\textbf{z}_k \in s_i} \chi \left ( {\delta  }_a - \min _ m f(\textbf{z}_k; \textbf{w}_m^t) \right ) ,
    \label{eq:perceptual loss}
    \end{equation}
where $\textbf{z}_k = \left [ x_k,~y_k,~1 \right ]$ is the homogeneous coordinate vector of the $k$-th point on $s_i$, $\textbf{z} \in t_m$ implies the following constraint $ \textbf{z}^T \textbf{w}_m^t\textbf{z} = 0 \left ( x_1 < x < x_2, y_1 < y < y_2 \right )$, $\left ( x_1,~y_1 \right )$ and  $\left ( x_2,~y_2 \right )$ are the starting and end points of $t_m$, and $\textbf{w}_m^t$ is the parameter matrix of the $m$-th curve $t_m$ in $\mathcal{T}$. $\chi(x)$ represents a step function. When $x > 0$ , $\chi(x) = 1$; otherwise, $\chi(x) = 0$. $K_{i}$ denotes the total number of points on $s_{i}$, and ${\delta }_a (=4)$ is a distance threshold. If $\sigma _{i} > {\delta }_b ( = 0.8)$, the curve $s_{i}$ is added into the vectorized structure set $\mathcal{E} = \left \{ e_{i} \right \}^{N_{\mathcal{E}}}_{i=1}$.
Finally, pixels on or near the curves in $\mathcal{E}$ are labeled as structural pixels, and a binary mask can be generated for them.

\section{Local Statistical Rectification}
As shown in Figure~\ref{Fig:fusion and synthesis}, although the fusion step in the previous section produces results with clear structures and natural stochastic details, it can still be further enhanced. For example, a minor stochastic component could be added onto the structural pixels in the fused image. In addition, since GANs are unsupervised, they cannot make a synthesized stochastic component have prescribed local statistical properties.
To make the synthesized results more natural looking and eliminate deviations introduced into the stochastic component in earlier stages, we propose a local statistical rectification step, which matches the local statistics of the synthesized image with the same local statistics of the groundtruth high resolution image. We discover that the combination of a local Gram matrix loss and a local correlation matrix loss can achieve this goal.

\subsection{Local Gram Matrix} The feature map, $\boldsymbol{F}$, generated by a convolutional layer can be computed once we run a convolutional neural network over an image $\boldsymbol{\mathcal{I}}$. For the location $(m, n)$ in the feature map, in its $r \times r$ neighborhood, we calculate a local Gram matrix $G^l(m, n) \in \mathbb{R}^{C_l \times C_l}$ in a similar manner as in~\cite{gatys2015texture} except that the Gram matrix is computed for a square window centered at the location instead of over the entire feature map. Here $C_l$ is the number of channels in the feature map generated by the $l$-th convolutional layer. The element $G^l_{ij}(m, n)$ in the local Gram matrix represents the correlation between the $i$-th channel and the $j$-th channel of the $l$-th layer in the $r \times r$ neighborhood of the location.
    \begin{equation}
      G_{ij}^l(m, n) = \frac{1}{r^2} \sum_{k \in R_{mn}} F_{ik}^l F_{jk}^l,
    \label{eq:gram matrix}
    \end{equation}
where $R_{mn}$ is the $r \times r$ neighborhood centered at $(m,n)$, and $F_{ik}^l$ is the value of the $i$-th channel of the feature map generated by the $l$-th convolutional layer at location $k \in R_{mn}$. Let $\boldsymbol{\mathcal{I}}$ and $\widehat{\boldsymbol{\mathcal{I}}}$ be the groundtruth image and the synthesized image, and $G^l$ and $\widehat{G}^l$ be their respective collections of local Gram matrices computed for the $l$-th layer. Then the local Gram matrix loss (L2 norm) for the $l$-th layer is calculated as follows,
    \begin{equation}
        \mathcal{L}^l_{Gram} = \frac{1}{2 {C_l}{M_l} N_l} \sum_{m,n} \sum_{{i,j}} \left\| G^l_{ij}(m, n) - \widehat{G}^l_{ij}(m, n) \right\|_2.
    \label{eq:local gram matrix}
    \end{equation}

\subsection{Local Correlation Matrix} The Gram matrix describes the degree of correlation between different feature channels. Here we also incorporate spatial correlation matrixes to describe the stochastic component over local regions. For the location $(m,n)$ in the $l$-th convolutional layer, the local correlation matrix $R^{l,c} \in \mathbb{R}^{M \times N}$ for a neighborhood of $(m,n)$ and the $c$-th feature channel is defined as follows,
    \begin{equation}
      R_{ij}^{l,c}(m, n) = \frac{1}{r^2} \sum_{p,q \in R_{mn}}  w_{ij}^{mn} f_{p,q}^{l,c} f_{p-i,q-j}^{l,c},
    \label{eq:gram matrix}
    \end{equation}
where $f_{p,q}^{l,c}$ is the value of the $c$-th feature channel of the $l$-th convolutional layer at location $(p, q)$, $i \in \left [ -\frac{r}{2}, \frac{r}{2} \right ]$ and $j \in \left [ -\frac{r}{2}, \frac{r}{2} \right ]$, and $(p-i,q-j) \in R_{mn}$. $R_{ij}^{l,k}(m,n)$ represents shifting the local feature patch $f^{l,k}_{m,n}$ by $i$ locations vertically and $j$ locations horizontally and then performing a point-wise multiplication between the original and shifted patches. $w^{mn}_{ij}$ is the inverse of the area of the overlapping region between these two patches, and is written as
    \begin{equation}
     w^{mn}_{ij} = \left[ \left( r - \left| i \right| \right) \left( r - \left| j \right| \right) \right]^{-1}.
    \label{weights}
    \end{equation}

Then the local correlation matrix loss for the $l$-th layer can be written as follows.
    \begin{equation}
       \mathcal{L}^l_{corr} = \frac{1}{2 {C_l}{M_l} N_l} \sum_{k} \sum_{m,n} \sum_{i,j} \left\| R^{l,k}_{ij}(m, n) - \widehat{R}^{l,k}_{ij}(m, n) \right\|_2.
    \label{eq:local gram matrix}
    \end{equation}

\subsection{Combined Loss} We use the same network architecture for both image fusion and local statistical rectification. We find out that the loss for image fusion still needs to be incorporated when local statistical rectification is performed. Thus, the loss for local statistical rectification is a combination of the loss for image fusion, the local Gram matrix loss and the local correlation matrix loss. The combined loss $\mathcal{L}_t$ is defined as follows.
    \begin{equation}
        \mathcal{L}_{s} =  \sum_l \left( \beta _0 ^l \mathcal{L}^l_{Gram} + \beta_1 ^l \mathcal{L}^l_{corr}  \right) + \mathcal{L}_{greg},
    \label{eq:texture_loss}
    \end{equation}
where ($\beta_0^l$, $\beta_1^l$) are weights for the local Gram matrix loss and local correlation matrix loss respectively.

We use the VGG-16 network~\cite{simonyan2014very} pre-trained on ImageNet ILSVRC 2012 training set~\cite{krizhevsky2012imagenet} to compute the feature maps used in the local Gram matrix loss and local correlation matrix loss. Denote layers "Pool1", "Pool2", "Pool3" and "Conv4\_3+ReLU" as the 1-st, 2-nd, 3-rd and 4-th layer, respectively. As shown in Figure~\ref{Fig:synthesis network}, the local correlation matrix loss is only defined over the feature map generated by the layer "Pool2", and the local Gram matrix loss is defined over the feature maps generated by layers "Pool3" and "Conv4\_3+ReLU", respectively. $\beta_1 ^2$ is set to 1e-11, both $\beta_0 ^3$ and $\beta_0 ^4$ are set to 1e-10.
The weights of individual losses are set to keep their gradients at the same order of magnitude. Neighborhood size is set to $7 \times 7$, $ 5 \times 5$ and $3 \times 3$ for $\mathcal{L} _{corr} ^2$, $\mathcal{L} _{Gram}^3$ and $\mathcal{L} _{Gram}^4$, respectively.

\section{Experimental Results}
Both the high-resolution deterministic component reconstruction and image fusion networks are implemented using Caffe~\cite{jia2014caffe}. GAN-based high resolution stochastic component synthesis is implemented using Tensorflow on the basis of the code from TensorLayer\footnote{\url{https://github.com/tensorlayer/srgan}}. All experiments run on NVIDIA TITAN X (Maxwell) GPUs with 12GB memory. Network training for deterministic component reconstruction takes 1 week using 4 GPUs in parallel while training for image fusion and local statistical rectification takes two days on 1 GPU. During testing, it takes around 135 seconds to generate the final result when the input is 270 x 480, and the output is 1080 x 1920. Generating the deterministic component only takes 60 seconds.

\begin{table*}[t]\small
	\centering
	\caption{Comparison of numerical results on Set5~\cite{bevilacqua2012low}, Set14~\cite{zeyde2010single}, B100~\cite{martin2001database}, Urban100~\cite{huang2015single} and the DIV2K validation set~\cite{Timofte_2017_CVPR_Workshops}. Average PSNR/SSIM values for scaling factors $\times$ 2, $\times$ 3 and $\times$ 4 are reported. The best results are shown in red, and the second best are shown in blue except for the ablation study results.}
	\label{benchmark results}
	\resizebox{1\textwidth}{!}
	{
		\begin{tabular}{c|c|c|c|c|c|c|c|c|c|c|c}
			\hline
			\hline
			Dataset        & Scale          & Bicubic      & SRCNN        & LapSRN       & DRRN         & MemNet       & EDSR         & RDN (\textbf{Baseline})    & w/o GGrad      & w/o Orientation    & DTSN(Ours)          \\ \hline
			& $\times$ 2     & 33.66/0.9299 & 36.66/0.9542 & 37.52/0.9591 & 37.74/0.9591 & 37.78/0.9597 & 38.11/0.9601 & {\color{red} 38.24/0.9614} & 38.23/0.9611   & 38.23/0.9611 & {\color{blue} 38.23/0.9611}  \\
			Set5           & $\times$ 3     & 30.39/0.8682 & 32.75/0.9090 & 33.82/0.9227 & 34.03/0.9244 & 34.09/0.9248 & 34.65/0.9282 & {\color{blue}34.71/0.9296} & 34.72/0.9298   & 34.73/0.9299 & {\color{red} 34.73/0.9299}  \\
			& $\times$ 4     & 28.42/0.8104 & 30.48/0.8628 & 31.54/0.8855 & 31.68/0.8888 & 31.74/0.8893 & 32.46/0.8968 & {\color{blue}32.47/0.8990} & 32.55/0.8988   & 32.57/0.8995 & {\color{red}32.58/0.9001}  \\ \hline
			& $\times$ 2     & 30.24/0.8688 & 32.45/0.9067 & 33.23/0.9130 & 33.23/0.9136 & 33.28/0.9142 & 33.92/0.9195 & {\color{blue}34.01/0.9212} & 34.10/0.9216   & 34.09/0.9216 & {\color{red}34.10/0.9217}  \\
			Set14          & $\times$ 3     & 27.55/0.7742 & 29.30/0.8215 & 29.79/0.8320 & 29.96/0.8349 & 30.00/0.8350 & 30.52/0.8462 & {\color{blue}30.57/0.8468} & 30.59/0.8469   & 30.58/0.8469 & {\color{red}30.59/0.8469}  \\
			& $\times$ 4     & 26.00/0.7027 & 27.50/0.7513 & 28.19/0.7720 & 28.21/0.7721 & 28.26/0.7723 & 28.80/0.7876 & {\color{blue}28.81/0.7871} & 28.87/0.7878   & 28.84/0.7875 & {\color{red}28.89/0.7882}  \\ \hline
			& $\times$ 2     & 29.56/0.8431 & 31.80/0.8950 & 31.80/0.8950 & 32.05/0.8973 & 32.08/0.8978 & 32.32/0.9013 & {\color{blue}32.34/0.9017} & 32.36/0.9020   & 32.36/0.9020 & {\color{red}32.36/0.9020}  \\
			B100           & $\times$ 3     & 27.21/0.7385 & 28.82/0.7973 & 28.82/0.7973 & 28.95/0.8004 & 28.96/0.8001 & 29.25/0.8093 & {\color{blue}29.26/0.8093} & 29.29/0.8095   & 29.29/0.8095 & {\color{red}29.29/0.8095}  \\
			& $\times$ 4     & 25.96/0.6675 & 27.32/0.7280 & 27.32/0.7280 & 27.38/0.7284 & 27.40/0.7281 & 27.71/0.7420 & {\color{blue}27.72/0.7419} & 27.73/0.7421   & 27.73/0.7421 & {\color{red}27.74/0.7421}  \\ \hline
			Urban100       & $\times$ 2     & 26.88/0.8403 & 30.41/0.9101 & 30.41/0.9101 & 31.23/0.9188 & 31.31/0.9195 & {\color{blue}32.93/0.9351} & 32.89/0.9353 & 32.96/0.9356   & 32.96/0.9356 & {\color{red}32.96/0.9356}  \\
			& $\times$ 4     & 23.14/0.6577 & 25.21/0.7553 & 25.21/0.7553 & 25.44/0.7638 & 25.50/0.7630 & {\color{blue}26.64/0.8033} & 26.61/0.8028 & 26.66/0.8038   & 26.65/0.8035 & {\color{red}26.70/0.8040}  \\ \hline
			& $\times$ 2     & 31.35/0.9076 & -/-          & -/-          & -/-          & -/-          & 35.03/0.9475 & {\color{blue}35.17/0.9478} & 35.26/0.9483   & 35.26/0.9483 & {\color{red}35.26/0.9483}  \\
			DIV2K          & $\times$ 3     & 28.49/0.8339 & -/-          & -/-          & -/-          & -/-          & 31.26/0.8910 & {\color{blue}31.39/0.8919} & 31.49/0.8931   & 31.48/0.8928 & {\color{red}31.49/0.8930}  \\
			& $\times$ 4     & 26.92/0.7774 & -/-          & -/-          & -/-          & -/-          & 29.25/0.8440 & {\color{blue}29.34/0.8442} & 29.44/0.8446   & 29.43/0.8443 & {\color{red}29.46/0.8451}  \\ \hline
			\hline
		\end{tabular}
	}
\end{table*}

\begin{figure*}[t]
  \centering
  \includegraphics[width=1.0\linewidth]{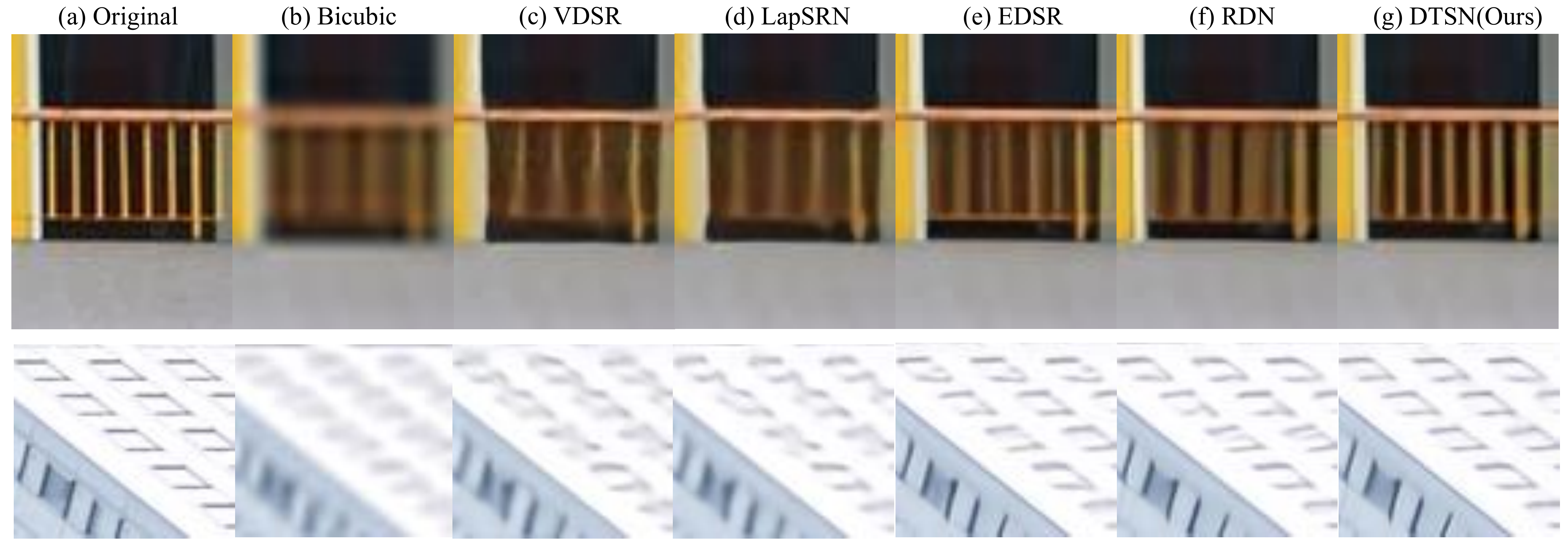}
  \caption{Visual comparison of deterministic component reconstruction results for the $\times 4$ scale. The input images are taken from the DIV2K~\cite{Timofte_2017_CVPR_Workshops} validation set.}
  \label{Fig:structural visualization}
\end{figure*}


\begin{figure*}[t]
  \centering
  \includegraphics[width=0.85\linewidth]{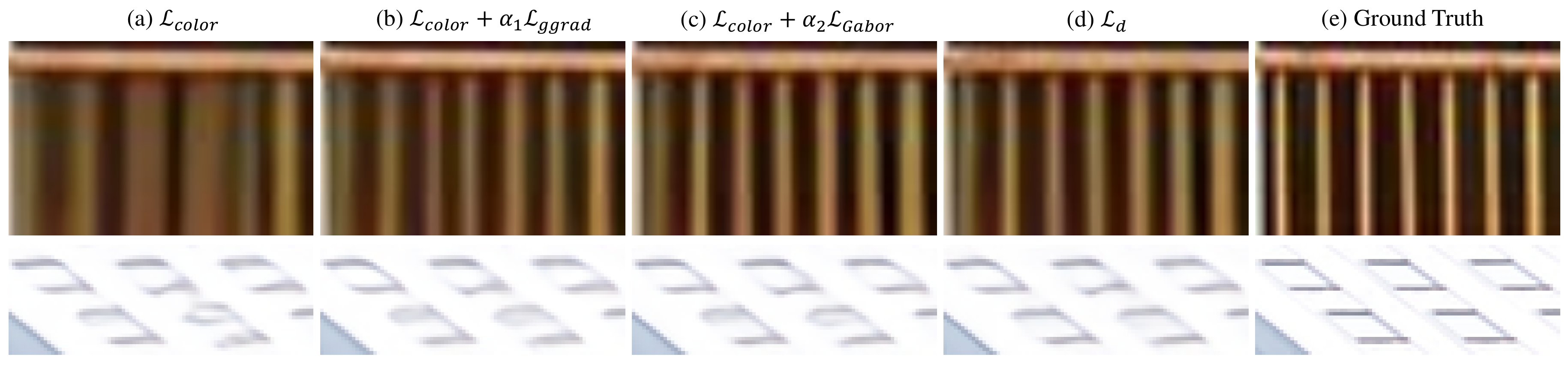}
  \caption{Effects of individual losses on deterministic component reconstruction for the $\times 4$ scale. $\mathcal{L}_{color}$ represents the results from RDN~\cite{zhang2018residual}.}
  \label{Fig:structural visualization2}
\end{figure*}

\begin{figure*}[t]
  \centering
  \includegraphics[width=1.0\linewidth]{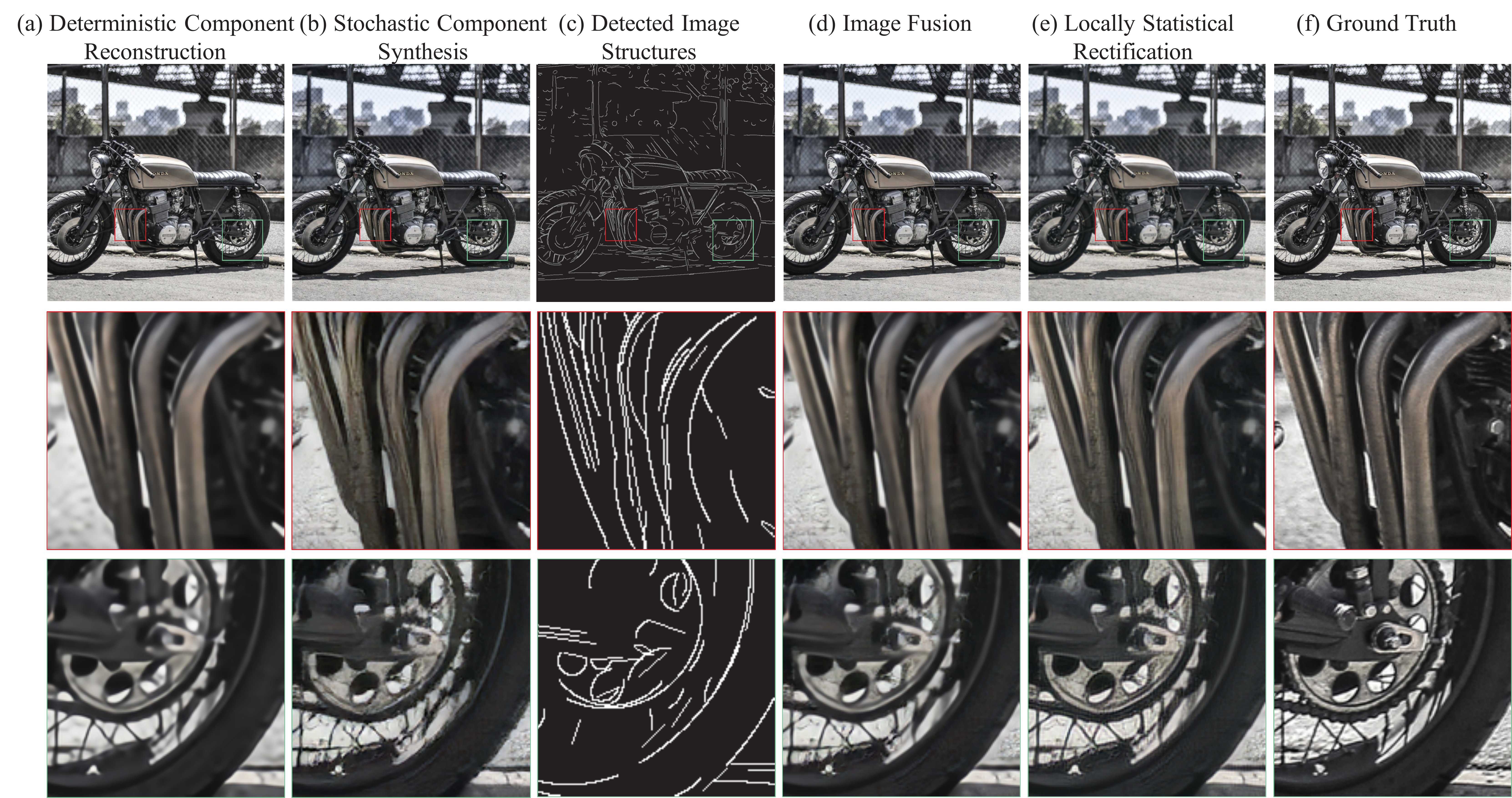}
  \caption{Visual comparison of intermediate results from our complete image superresolution pipeline for the "$4\times$" scale.}
  \label{Fig:fusion and synthesis}
\end{figure*}

\begin{figure*}[t]
\setlength{\abovecaptionskip}{1pt}
\setlength{\belowcaptionskip}{0pt}
  \centering
  \includegraphics[width=1.0\linewidth]{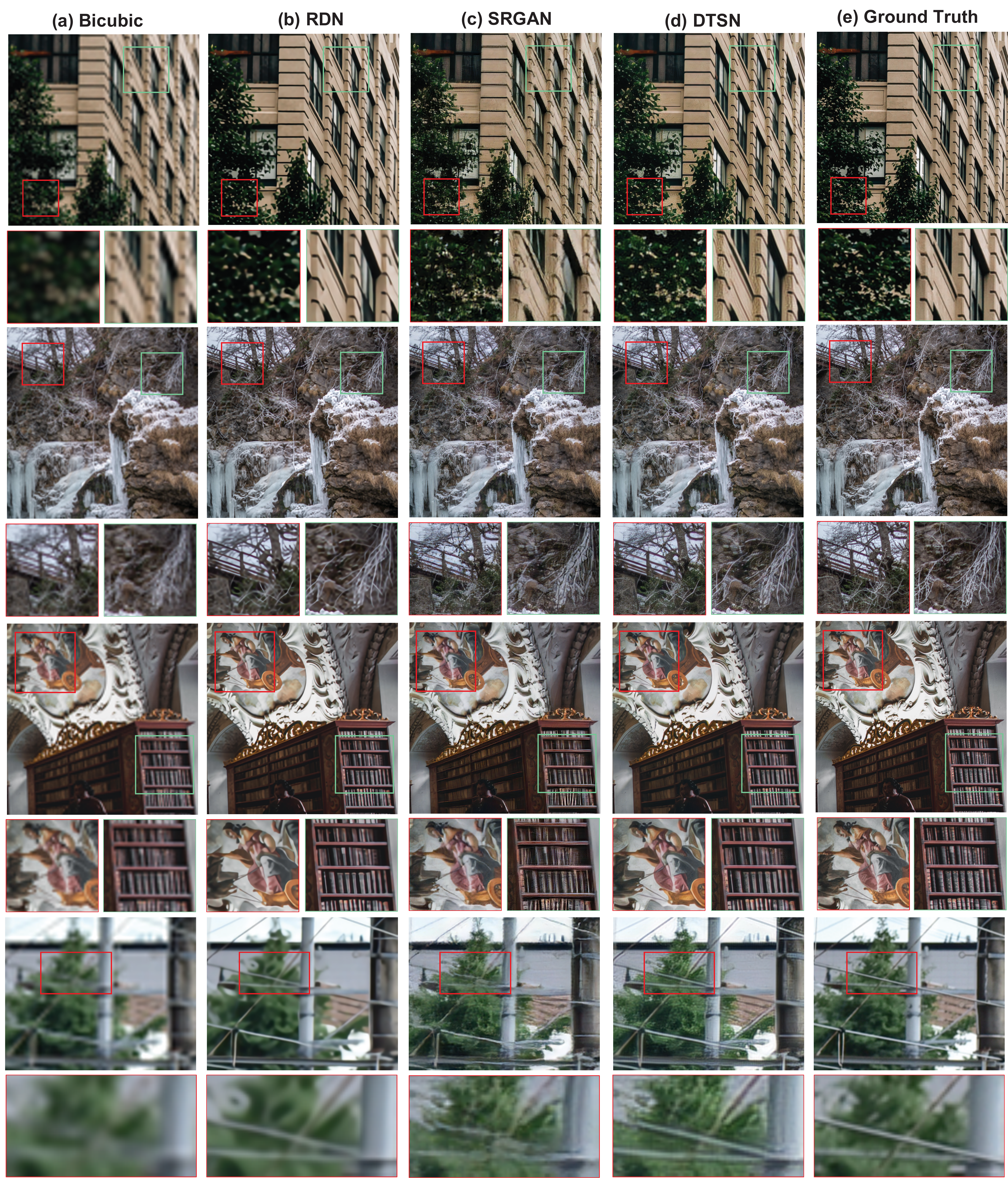}
  \caption{Visual comparison of final superresolution results from our method (DTSN) and other state-of-the-art algorithms for the $\times 4$ scale. The input images are taken from the DIV2K~\cite{Timofte_2017_CVPR_Workshops} validation set. More results are given in the supplementary materials.}
  \label{Fig:dtsn visualization}
\end{figure*}

\subsection{High-Resolution Deterministic Component Reconstruction}
\subsubsection{Network Settings and Parameters}
In the network for deterministic component reconstruction, $\phi_{s}\left ( \cdot,{\theta}_s \right )$, the kernel size of all the convolutional layers except the $1 \times 1$ convolutional layer after the concatenation layer ("Concat" in Figure~ \ref{Fig:image structure}) is $3 \times 3$. In Figure~\ref{Fig:image structure}(b), the number of channels in Conv1 through Conv4 is 64. There are 8 convolutional layers in a residual dense block, and a total of 16 residual dense blocks in $\phi_{s}\left ( \cdot,{\theta}_s \right )$. In the pixel shuffle module, the number of channels in Conv1 is 256, 576 and 256 for "$2 \times$", "$3 \times$" and "$4 \times$" superresolution, respectively. For "$4 \times$" superresolution, the Conv2 layer has 256 channels. The final output layer Conv5 has 3 channels. The stride for all the convolutional layers is 1. We use ReLU as the nonlinear activation function following all convolutional layers.

\subsubsection{Datasets and Performance Evaluation}
In the NTIRE 2017 Challenge on Single Image Super Resolution, Timofte {\em et al.}~\cite{Timofte_2017_CVPR_Workshops} released a high-quality (2K resolution) dataset, called DIV2K, which consists of 800 training images, 100 validation images, and 100 test images. It has become a standard benchmark for single image super resolution. In our experiments, we train our deep network on the training set of DIV2K, and test it on five standard benchmark datasets including Set5~\cite{bevilacqua2012low}, Set14~\cite{zeyde2010single}, B100~\cite{martin2001database}, Urban100~\cite{huang2015single} and DIV2K validation set~\cite{Timofte_2017_CVPR_Workshops}. Note that the ground truth of the DIV2K test set is not publicly available. Therefore, we use the DIV2K validation set during testing. Images in this validation set were never used for training or parameter tuning. PSNR and SSIM~\cite{wang2004image} of the Y channel (\emph{i.e.}, luminance) of the transformed YCbCr space are used to quantitatively evaluate the performance of different SR algorithms in addition to user studies.

\subsubsection{Training Details}
During network training, each mini-batch contains 5 training samples. The low-resolution input in each training sample is a $36 \times 36$ RGB patch. Following the data augmentation setting in~\cite{lim2017enhanced}, every patch is filpped horizontally or vertically and then rotated by $90^{\circ}$ randomly. We train the network with the Adam optimizer. The initial learning rate is set to 0.0001, and the weight decay is set to 0.0005. Due to large on-board memory consumption, each experiment is run in parallel on 4 TITAN X (Maxwell) GPUs. Training converges after 350000 iterations. The learning rate is decreased by half every 70000 iterations.

\subsubsection{Comparisons}
We have compared our high-resolution deterministic component reconstruction algorithm with six existing state-of-the-art superresolution algorithms, including SRCNN~\cite{dong2016image}, LapSRN~\cite{lai2017deep}, DRRN~\cite{tai2017image}, MemNet~\cite{tai2017memnet}, EDSR~\cite{lim2017enhanced}, and RDN~\cite{zhang2018residual}. Table~\ref{benchmark results} compares PSNR/SSIM results for three scales ($\times$ 2, $\times$ 3, and $\times$ 4) on five different benchmarks, including Set5~\cite{bevilacqua2012low}, Set14~\cite{zeyde2010single}, B100~\cite{martin2001database}, Urban100~\cite{huang2015single} and DIV2K validation set~\cite{Timofte_2017_CVPR_Workshops}. All the six existing SISR algorithms participating in this comparison are deep learning based methods. They treat the SISR problem as a regression problem. The loss used their network training is the mean color loss with the $L_1$ norm (mean absolute error, MAE) or $L_2$ norm (mean squared error, MSE). From SRCNN to RDN, the performance (PSNR/SSIM) of the regression results improves due to better network architectures and training strategies. In addition to such advances, we introduce two new loss functions, the generalized gradient loss and the orientation loss, to better preserve curvilinear structures. Experiments show that our algorithm achieves the best performance on all the benchmarks for all the scales except the "$2 \times$" scale on Set5. When compared with state-of-the-art algorithms, the degree of improvement (~0.1dB) on a saturated benchmark such as DIV2K is actually substantial. In comparison to RDN~\cite{zhang2018residual}, which is the previous best-performing algorithm, the proposed loss functions further improve PSNR and SSIM. In particular, on the popular DIV2K validation set~\cite{Timofte_2017_CVPR_Workshops}, our combined loss function outperforms the mean color loss alone by 0.09 db, 0.10 db and 0.12 db on the "$2 \times$", "$3 \times$" and "$4 \times$" scales, respectively. Figure~\ref{Fig:structural visualization} compares visual results from different state-of-the-art algorithms. Our proposed algorithm exhibits a stronger ability than EDSR and RDN in preserving structural lines and curves.

\subsubsection{Ablation Study}
To verify the effectiveness of the generalized gradient loss and the orientation loss, we remove one of them from the combined loss every time, and use the rest of the terms in the loss to train the network while keeping all other settings exactly the same as before. The results for the "$2 \times$", "$3 \times$" and "$4 \times$" scales are also reported in Table~\ref{benchmark results}. When we remove the orientation loss, the performance drop ranges from 0.01 db to 0.05 db on the five benchmarks. When we remove the generalized gradient loss, the performance also drops similarly. This indicates that both the generalized gradient loss and the orientation loss can improve the regression accuracy. When they are combined, a larger performance gain can be achieved.

\subsection{High-Resolution Stochastic Component Hallucination}
\subsubsection{Network Settings and Parameters}
We use the same network settings as in the original paper~\cite{Ledig_2017_CVPR} in our implementation. There are 16 residual blocks in the generator network. The kernel size of all the convolutional layers except the input and output layers is $3 \times 3$. The number of channels in all the convolutional layers except the output layer is 64. The kernel size of the input and output layers is $9 \times 9$ and the number of channels in the output layer is 3. The discriminator network is a VGG-type network that has 8 convolutional layers including the input layer. The kernel size of all the convolutional layers is $3 \times 3$. The number of channels increases by a factor of 2 every time we move from one layer to the next. Like in the VGG network, the number of channels in a convolutional layer increases from 64 to 512. "Dense 1" has 1024 outputs and "Dense 2" has 1 output.

\subsubsection{Training Details}
We train the network on 115 thousand images from the ImageNet dataset and the RAISE dataset. Each mini-batch has 8 training samples, and the low-resolution input in each sample is a $48 \times 48$ RGB patch. We train the network with the Adam optimizer ($\beta=0.9$). The initial learning rate is set to 0.0001. Experiments related to this network run on a single TITAN X (Maxwell) GPU. Training converges after 1000000 iterations. The learning rate is decreased by a factor of ten every 400000 iterations. We first train the generator alone until convergence, and then alternate the training of the generator and discriminator.

\subsection{Image Fusion and Local Statistical Rectification}
\subsubsection{Network Settings and Parameters}
There are three inputs to the image fusion and local statistics rectification network shown in Figure~\ref{Fig:synthesis network}. All convolutional layers except those in the upscaling module, the residual blocks and the final output layer, use $3 \times 3$ kernels and have 64 channels. The upscaling module is the same as that in $\phi_{s}\left ( \cdot,{\theta}_s \right )$. In a residual block, the Conv1 layer has 64 channels and uses $1 \times 1$ kernels; the Conv2 layer has 64 channels and uses $3 \times 3$ kernels. The output layer Conv12 has 3 channels. The stride is set to $1$ for all convolutional layers. PReLU is used as the activation function following all convolutional layers.

\subsubsection{Datasets and Training Details}
We train the image fusion and local statistics rectification network on the DIV2K training set and test its performance on the DIV2K validation set as well as other benchmarks. We train the image fusion network $\phi_{t}\left ( \cdot,{\theta}_t \right )$ using the guided regression loss $\mathcal{L}_{greg}$ at first. The initial learning rate is set to 0.0001, and the weight decay is set to 0.0005. The network is trained using a single TITAN X (Maxwell) GPU. Training stops after 90000 iterations. The learning rate is decreased by half every 30000 iterations. Then we continue to train the local statistics rectification network using the combined loss function $\mathcal{L}_{t}$. The learning rate is kept at 0.00001 at the beginning, and is decreased by half every 10000 iterations. The training stops after 30000 iterations. The Adam optimizer is used as in the previous stages.

\subsubsection{Results}
Figure~\ref{Fig:fusion and synthesis} shows a visual comparison of intermediate results in our complete image superresolution pipeline for the "$4 \times$" scale. Figure~\ref{Fig:fusion and synthesis}(a) shows the high resolution deterministic component reconstruction result. Some non-structural pixels are blurred, but high-resolution edges are well reconstructed. Figure~\ref{Fig:fusion and synthesis}(b) shows GAN-based stochastic component synthesis result. GAN clearly adds more realistic stochastic details, but gives rise to edge aliasing at the same time. Figure~\ref{Fig:fusion and synthesis}(c) shows detected image structures. Figure~\ref{Fig:fusion and synthesis}(d) indicates that reconstructed deterministic components and synthesized stochastic components can be effectively fused together using the guided regression loss. After comparing the images in Figure~\ref{Fig:fusion and synthesis}(d) and (e), we can find that local statistics rectification effectively enforces local statistical properties by refining previously synthesized stochastic components to make the fused results have more natural and realistic details. Compared with the ground truth in Figure~\ref{Fig:fusion and synthesis}(f), Figure~\ref{Fig:fusion and synthesis}(e) presents very competitive results.

In Figure~\ref{Fig:dtsn visualization}, we compare our final superresolution results (DTSN) with those generated from state-of-the-art SISR algorithms (RDN \cite{zhang2018residual} and SRGAN~\cite{Ledig_2017_CVPR}). Each odd numbered row shows the raw results from different algorithms. Magnified local image patches from such results are shown in the row immediately below. These results show that RDN can produce high-quality edges and curves, but fail to generate fine stochastic details for non-structural pixels. SRGAN can generate natural stochastic details for all pixels without maintaining a proper balance between the stochastic and deterministic components. The results from the proposed superresolution algorithm show abundant stochastic details, which look more natural and pleasing than SRGAN, while exhibiting high-quality curvilinear structures, such as curves and lines. Figure~\ref{Fig:dtsn visualization} demonstrates that the proposed DTSN produces more natural and accurate high-resolution images than both EDSR and SRGAN, and these images are appealing even when compared with the ground truth.

\subsection{User Study}
As discussed earlier, with respect to a downsampled low resolution image, a high resolution image can be decomposed into two components, a deterministic component and a stochastic component. The deterministic component can be recovered from the low-frequency signals in the downsampled image by exploiting the correlation between the two. The stochastic component, on the other hand, contains the rest of the signals in the high resolution image that have little correlation with the low resolution image. By using the low resolution image as an anchor, the reconstructed high resolution deterministic component is fully registered with the groundtruth high resolution image. Therefore, any quality measures that rely on pixelwise correspondences, such as PSNR, could be used to evaluate the reconstructed results.

On the other hand, the high resolution stochastic component cannot be uniquely recovered from the low resolution image. We can only "hallucinate" this stochastic component through various image synthesis techniques, including GANs and statistical property matching. The synthesized stochastic component may be either statistically or perceptually similar to the stochastic details in the ground truth, but does not have strict pixelwise correspondences with the latter. Therefore, quality measures based on pixel-level registration is not well suited for evaluating image superresolution results with stochastic components. Instead, subjective evaluation based on user studies is a much more appropriate means to evaluate our final superresolution results.

To test the perceptual quality of different methods, we developed an online user study, which asks participants to mark superresolution images generated by SRGAN, RDN, EDSR and our method. We cropped 25 images from the DIV2K validation set in order to make sure that the group of output images from all methods can be displayed on the same screen and they contain recognizable foreground objects. All generated images are 4x larger than the input and divided into 25 groups. Each group has 4 images from different methods using the same input. Thus there are a total of $25 \times 4$ images.


Every participant is asked to assign integral marks to all images within 10 randomly chosen groups. The minimum score is 1, which corresponds to very poor perceptual quality. And the maximum score is 10, which corresponds to very high perceptual quality. The display order of 4 images within the same group is randomized for every group and participant. There were 42 worldwide participants and 6 of them claimed they were photographers. A brief tutorial was given by the online study before they started the test. We encouraged participants to spend at least 3 minutes on each group.


At the end, for each method, we collected 420 scores. The mean score of our method ($\mu_b =7.59$) is the highest among all 4 methods. To verify if our mean score is significantly higher than other methods'($\mu_a$), a paired T-test was performed between our method and every other method with the following hypotheses:
    \begin{align*}
        H_0 &: \mu_a \ge \mu_b, \\
        H_1 &: \mu_a < \mu_b.
    \label{eq:hypotheses}
    \end{align*}

Hypothesis $H_1$ represents the mean score of our method is higher than the compared method. We test those hypotheses against SRGAN, RDN and EDSR. The results are shown in Table \ref{tbl:ttest}.

\begin{table}[]
\centering
\caption{Paired T-test Results ($\alpha=0.05$)}
\label{tbl:ttest}
\resizebox{0.25\textwidth}{!}
{
\begin{tabular}{cccc}
\hline
Method & $\mu_a$  & T      & P(two-tail) \\ \hline
SRGAN  & $5.69$   & -15.67 & 7.43e-44    \\ \hline
RDN    & $4.76$   & -23.01 & 2.47e-76    \\ \hline
EDSR   & $4.75$   & -23.58 & 7.60e-79    \\ \hline
\end{tabular}
}
\end{table}

Note that all P values are less than 0.05 and all T values are negative, which means we can reject $H_0$ and accept $H_1$ with statistical significance. This gives rise to the conclusion that our method is capable of generating images that is significantly perceptually better.

\subsection{Limitations}
Our proposed method has the following limitations. First, its performance heavily relies on the accuracy of the structure detection algorithm. ELSD can only detect line segments, circular and elliptical arcs. When there exist more complex curves, their detected location may drift around the true location. Second, We have found that the local Gram matrix loss may cause halos/ringing around certain strong edges. When the local Gram matrix loss is only added to deeper layers which are farther from the input layer, the halo/ringing effect is likely to be reduced. Third, the hallucinated details during local statistical rectification are too strong in certain images. We have experimentally found that decreasing the weight of the local Gram matrix loss and the local correlation matrix loss can potentially make the hallucinated details less noticeable.

\section{Conclusions}
In this paper, we have presented a new pipeline for single image superresolution. Different from previous algorithms, we model a single pixel using two complementary components -- a deterministic component and a stochastic component. The SISR problem is decomposed into independent problems generating these components. Then we fuse the two components using a deep neural network that also performs local statistical rectification, which tries to make the local statistics of the fused image match the same local statistics of the groundtruth image. Quantitative results and a user study indicate that the proposed method outperforms existing state-of-the-art algorithms with a clear margin.

\section*{Acknowledgments}
We would like to thank the anonymous reviewers for their valuable comments. This work was partially supported by Hong Kong Research Grants Council under General Research Funds (HKU17209714).

\bibliographystyle{ACM-Reference-Format}
\bibliography{superres}


\begin{thebibliography}{42}


\ifx \showCODEN    \undefined \def \showCODEN     #1{\unskip}     \fi
\ifx \showDOI      \undefined \def \showDOI       #1{#1}\fi
\ifx \showISBNx    \undefined \def \showISBNx     #1{\unskip}     \fi
\ifx \showISBNxiii \undefined \def \showISBNxiii  #1{\unskip}     \fi
\ifx \showISSN     \undefined \def \showISSN      #1{\unskip}     \fi
\ifx \showLCCN     \undefined \def \showLCCN      #1{\unskip}     \fi
\ifx \shownote     \undefined \def \shownote      #1{#1}          \fi
\ifx \showarticletitle \undefined \def \showarticletitle #1{#1}   \fi
\ifx \showURL      \undefined \def \showURL       {\relax}        \fi
\providecommand\bibfield[2]{#2}
\providecommand\bibinfo[2]{#2}
\providecommand\natexlab[1]{#1}
\providecommand\showeprint[2][]{arXiv:#2}

\bibitem[\protect\citeauthoryear{Ahn and Nam}{Ahn and Nam}{2016}]%
        {ahn2016texture}
\bibfield{author}{\bibinfo{person}{Il~Jun Ahn} {and} \bibinfo{person}{Woo~Hyun
  Nam}.} \bibinfo{year}{2016}\natexlab{}.
\newblock \showarticletitle{Texture Enhancement via High-Resolution Style
  Transfer for Single-Image Super-Resolution}.
\newblock \bibinfo{journal}{\emph{arXiv preprint arXiv:1612.00085}}
  (\bibinfo{year}{2016}).
\newblock


\bibitem[\protect\citeauthoryear{Baker and Kanade}{Baker and Kanade}{2000}]%
        {BK2000}
\bibfield{author}{\bibinfo{person}{S. Baker} {and} \bibinfo{person}{T.
  Kanade}.} \bibinfo{year}{2000}\natexlab{}.
\newblock \showarticletitle{Hallucinating faces}. In
  \bibinfo{booktitle}{\emph{4th IEEE International Conference on Automatic Face
  and Gesture Recognition}}. \bibinfo{pages}{83--88}.
\newblock


\bibitem[\protect\citeauthoryear{Bengtsson, Gu, Viberg, and
  Lindstrom}{Bengtsson et~al\mbox{.}}{2012}]%
        {BGVL2012}
\bibfield{author}{\bibinfo{person}{T. Bengtsson}, \bibinfo{person}{I.Y-H. Gu},
  \bibinfo{person}{M. Viberg}, {and} \bibinfo{person}{K. Lindstrom}.}
  \bibinfo{year}{2012}\natexlab{}.
\newblock \showarticletitle{Regularized optimization for joint super-resolution
  and high dynamic range image reconstruction in a perceptually uniform
  domain}. In \bibinfo{booktitle}{\emph{IEEE International Conference on
  Acoustics, Speech and Signal Processing}}. \bibinfo{pages}{1097–1100}.
\newblock


\bibitem[\protect\citeauthoryear{Bevilacqua, Roumy, Guillemot, and
  Alberi-Morel}{Bevilacqua et~al\mbox{.}}{2012}]%
        {bevilacqua2012low}
\bibfield{author}{\bibinfo{person}{Marco Bevilacqua}, \bibinfo{person}{Aline
  Roumy}, \bibinfo{person}{Christine Guillemot}, {and}
  \bibinfo{person}{Marie~Line Alberi-Morel}.} \bibinfo{year}{2012}\natexlab{}.
\newblock \showarticletitle{Low-complexity single-image super-resolution based
  on nonnegative neighbor embedding}.
\newblock  (\bibinfo{year}{2012}).
\newblock


\bibitem[\protect\citeauthoryear{Capel and Zisserman}{Capel and
  Zisserman}{1998}]%
        {CZ1998}
\bibfield{author}{\bibinfo{person}{D. Capel} {and} \bibinfo{person}{A.
  Zisserman}.} \bibinfo{year}{1998}\natexlab{}.
\newblock \showarticletitle{Automated mosaicing with superresolution zoom}. In
  \bibinfo{booktitle}{\emph{IEEE Conference on Computer Vision and Pattern
  Recognition}}. \bibinfo{pages}{885–891}.
\newblock


\bibitem[\protect\citeauthoryear{Chang, Yeung, and Xiong}{Chang
  et~al\mbox{.}}{2004}]%
        {chang2004super}
\bibfield{author}{\bibinfo{person}{Hong Chang}, \bibinfo{person}{Dit-Yan
  Yeung}, {and} \bibinfo{person}{Yimin Xiong}.}
  \bibinfo{year}{2004}\natexlab{}.
\newblock \showarticletitle{Super-resolution through neighbor embedding}. In
  \bibinfo{booktitle}{\emph{Computer Vision and Pattern Recognition, 2004. CVPR
  2004. Proceedings of the 2004 IEEE Computer Society Conference on}},
  Vol.~\bibinfo{volume}{1}. IEEE, \bibinfo{pages}{I--I}.
\newblock


\bibitem[\protect\citeauthoryear{Dong, Loy, He, and Tang}{Dong
  et~al\mbox{.}}{2014}]%
        {dong2014learning}
\bibfield{author}{\bibinfo{person}{Chao Dong}, \bibinfo{person}{Chen~Change
  Loy}, \bibinfo{person}{Kaiming He}, {and} \bibinfo{person}{Xiaoou Tang}.}
  \bibinfo{year}{2014}\natexlab{}.
\newblock \showarticletitle{Learning a deep convolutional network for image
  super-resolution}. In \bibinfo{booktitle}{\emph{European Conference on
  Computer Vision}}. Springer, \bibinfo{pages}{184--199}.
\newblock


\bibitem[\protect\citeauthoryear{Dong, Loy, He, and Tang}{Dong
  et~al\mbox{.}}{2016}]%
        {dong2016image}
\bibfield{author}{\bibinfo{person}{Chao Dong}, \bibinfo{person}{Chen~Change
  Loy}, \bibinfo{person}{Kaiming He}, {and} \bibinfo{person}{Xiaoou Tang}.}
  \bibinfo{year}{2016}\natexlab{}.
\newblock \showarticletitle{Image super-resolution using deep convolutional
  networks}.
\newblock \bibinfo{journal}{\emph{IEEE transactions on pattern analysis and
  machine intelligence}} \bibinfo{volume}{38}, \bibinfo{number}{2}
  (\bibinfo{year}{2016}), \bibinfo{pages}{295--307}.
\newblock


\bibitem[\protect\citeauthoryear{Fattal}{Fattal}{2007}]%
        {fattal2007}
\bibfield{author}{\bibinfo{person}{Raanan Fattal}.}
  \bibinfo{year}{2007}\natexlab{}.
\newblock \showarticletitle{Image Upsampling via Imposed Edge Statistics}. In
  \bibinfo{booktitle}{\emph{ACM SIGGRAPH 2007 Papers}}
  \emph{(\bibinfo{series}{SIGGRAPH '07})}. \bibinfo{publisher}{ACM},
  \bibinfo{address}{New York, NY, USA}, Article \bibinfo{articleno}{95}.
\newblock
\urldef\tempurl%
\url{https://doi.org/10.1145/1275808.1276496}
\showDOI{\tempurl}


\bibitem[\protect\citeauthoryear{Freeman, Jones, and Pasztor}{Freeman
  et~al\mbox{.}}{2002}]%
        {freeman2002example}
\bibfield{author}{\bibinfo{person}{William~T Freeman},
  \bibinfo{person}{Thouis~R Jones}, {and} \bibinfo{person}{Egon~C Pasztor}.}
  \bibinfo{year}{2002}\natexlab{}.
\newblock \showarticletitle{Example-based super-resolution}.
\newblock \bibinfo{journal}{\emph{IEEE Computer graphics and Applications}}
  \bibinfo{volume}{22}, \bibinfo{number}{2} (\bibinfo{year}{2002}),
  \bibinfo{pages}{56--65}.
\newblock


\bibitem[\protect\citeauthoryear{Gatys, Ecker, and Bethge}{Gatys
  et~al\mbox{.}}{2015}]%
        {gatys2015texture}
\bibfield{author}{\bibinfo{person}{Leon Gatys}, \bibinfo{person}{Alexander~S
  Ecker}, {and} \bibinfo{person}{Matthias Bethge}.}
  \bibinfo{year}{2015}\natexlab{}.
\newblock \showarticletitle{Texture synthesis using convolutional neural
  networks}. In \bibinfo{booktitle}{\emph{Advances in Neural Information
  Processing Systems}}. \bibinfo{pages}{262--270}.
\newblock


\bibitem[\protect\citeauthoryear{Gatys, Ecker, and Bethge}{Gatys
  et~al\mbox{.}}{2016}]%
        {gatys2016image}
\bibfield{author}{\bibinfo{person}{Leon~A Gatys}, \bibinfo{person}{Alexander~S
  Ecker}, {and} \bibinfo{person}{Matthias Bethge}.}
  \bibinfo{year}{2016}\natexlab{}.
\newblock \showarticletitle{Image style transfer using convolutional neural
  networks}. In \bibinfo{booktitle}{\emph{Proceedings of the IEEE Conference on
  Computer Vision and Pattern Recognition}}. \bibinfo{pages}{2414--2423}.
\newblock


\bibitem[\protect\citeauthoryear{Goodfellow, Pouget-Abadie, Mirza, Xu,
  Warde-Farley, Ozair, Courville, and Bengio}{Goodfellow et~al\mbox{.}}{2014}]%
        {goodfellow2014generative}
\bibfield{author}{\bibinfo{person}{Ian Goodfellow}, \bibinfo{person}{Jean
  Pouget-Abadie}, \bibinfo{person}{Mehdi Mirza}, \bibinfo{person}{Bing Xu},
  \bibinfo{person}{David Warde-Farley}, \bibinfo{person}{Sherjil Ozair},
  \bibinfo{person}{Aaron Courville}, {and} \bibinfo{person}{Yoshua Bengio}.}
  \bibinfo{year}{2014}\natexlab{}.
\newblock \showarticletitle{Generative adversarial nets}. In
  \bibinfo{booktitle}{\emph{Advances in neural information processing
  systems}}. \bibinfo{pages}{2672--2680}.
\newblock


\bibitem[\protect\citeauthoryear{Gunturk, Altunbasak, and Mersereau}{Gunturk
  et~al\mbox{.}}{2004}]%
        {GAM2004}
\bibfield{author}{\bibinfo{person}{B.K. Gunturk}, \bibinfo{person}{Y.
  Altunbasak}, {and} \bibinfo{person}{R.M. Mersereau}.}
  \bibinfo{year}{2004}\natexlab{}.
\newblock \showarticletitle{Superresolution reconstruction of compressed video
  using transform domain statistics}.
\newblock \bibinfo{journal}{\emph{IEEE Transactions on Image Processing}}
  \bibinfo{volume}{13}, \bibinfo{number}{1} (\bibinfo{year}{2004}),
  \bibinfo{pages}{33--43}.
\newblock


\bibitem[\protect\citeauthoryear{He, Zhang, Ren, and Sun}{He
  et~al\mbox{.}}{2015}]%
        {he2015delving}
\bibfield{author}{\bibinfo{person}{Kaiming He}, \bibinfo{person}{Xiangyu
  Zhang}, \bibinfo{person}{Shaoqing Ren}, {and} \bibinfo{person}{Jian Sun}.}
  \bibinfo{year}{2015}\natexlab{}.
\newblock \showarticletitle{Delving deep into rectifiers: Surpassing
  human-level performance on imagenet classification}. In
  \bibinfo{booktitle}{\emph{Proceedings of the IEEE international conference on
  computer vision}}. \bibinfo{pages}{1026--1034}.
\newblock


\bibitem[\protect\citeauthoryear{Hu, Lam, Qiu, and Shen}{Hu
  et~al\mbox{.}}{2010}]%
        {HLQS2010}
\bibfield{author}{\bibinfo{person}{Y. Hu}, \bibinfo{person}{K.M. Lam},
  \bibinfo{person}{G. Qiu}, {and} \bibinfo{person}{T. Shen}.}
  \bibinfo{year}{2010}\natexlab{}.
\newblock \showarticletitle{From local pixel structure to global image
  super-resolution: a new face hallucination framework}.
\newblock \bibinfo{journal}{\emph{IEEE Transactions on Image Processing}}
  \bibinfo{volume}{20}, \bibinfo{number}{2} (\bibinfo{year}{2010}),
  \bibinfo{pages}{433--445}.
\newblock


\bibitem[\protect\citeauthoryear{Huang, Singh, and Ahuja}{Huang
  et~al\mbox{.}}{2015}]%
        {huang2015single}
\bibfield{author}{\bibinfo{person}{Jia-Bin Huang}, \bibinfo{person}{Abhishek
  Singh}, {and} \bibinfo{person}{Narendra Ahuja}.}
  \bibinfo{year}{2015}\natexlab{}.
\newblock \showarticletitle{Single image super-resolution from transformed
  self-exemplars}. In \bibinfo{booktitle}{\emph{Proceedings of the IEEE
  Conference on Computer Vision and Pattern Recognition}}.
  \bibinfo{pages}{5197--5206}.
\newblock


\bibitem[\protect\citeauthoryear{Jia, Shelhamer, Donahue, Karayev, Long,
  Girshick, Guadarrama, and Darrell}{Jia et~al\mbox{.}}{2014}]%
        {jia2014caffe}
\bibfield{author}{\bibinfo{person}{Yangqing Jia}, \bibinfo{person}{Evan
  Shelhamer}, \bibinfo{person}{Jeff Donahue}, \bibinfo{person}{Sergey Karayev},
  \bibinfo{person}{Jonathan Long}, \bibinfo{person}{Ross Girshick},
  \bibinfo{person}{Sergio Guadarrama}, {and} \bibinfo{person}{Trevor Darrell}.}
  \bibinfo{year}{2014}\natexlab{}.
\newblock \showarticletitle{Caffe: Convolutional Architecture for Fast Feature
  Embedding}.
\newblock \bibinfo{journal}{\emph{arXiv preprint arXiv:1408.5093}}
  (\bibinfo{year}{2014}).
\newblock


\bibitem[\protect\citeauthoryear{Jianchao, Wright, Huang, and Ma}{Jianchao
  et~al\mbox{.}}{2008}]%
        {jianchao2008image}
\bibfield{author}{\bibinfo{person}{Yang Jianchao}, \bibinfo{person}{John
  Wright}, \bibinfo{person}{Thomas Huang}, {and} \bibinfo{person}{Yi Ma}.}
  \bibinfo{year}{2008}\natexlab{}.
\newblock \showarticletitle{Image super-resolution as sparse representation of
  raw image patches}. In \bibinfo{booktitle}{\emph{Proc. IEEE Conf. on Computer
  Vision and Pattern Recognition}}. \bibinfo{pages}{1--8}.
\newblock


\bibitem[\protect\citeauthoryear{Kim, Kwon~Lee, and Mu~Lee}{Kim
  et~al\mbox{.}}{2016}]%
        {kim2016accurate}
\bibfield{author}{\bibinfo{person}{Jiwon Kim}, \bibinfo{person}{Jung Kwon~Lee},
  {and} \bibinfo{person}{Kyoung Mu~Lee}.} \bibinfo{year}{2016}\natexlab{}.
\newblock \showarticletitle{Accurate image super-resolution using very deep
  convolutional networks}. In \bibinfo{booktitle}{\emph{Proceedings of the IEEE
  Conference on Computer Vision and Pattern Recognition}}.
  \bibinfo{pages}{1646--1654}.
\newblock


\bibitem[\protect\citeauthoryear{Lai, Huang, Ahuja, and Yang}{Lai
  et~al\mbox{.}}{2017a}]%
        {LapSRN}
\bibfield{author}{\bibinfo{person}{Wei-Sheng Lai}, \bibinfo{person}{Jia-Bin
  Huang}, \bibinfo{person}{Narendra Ahuja}, {and} \bibinfo{person}{Ming-Hsuan
  Yang}.} \bibinfo{year}{2017}\natexlab{a}.
\newblock \showarticletitle{Deep Laplacian Pyramid Networks for Fast and
  Accurate Super-Resolution}. In \bibinfo{booktitle}{\emph{Proceedings of the
  IEEE Conference on Computer Vision and Pattern Recognition}}.
\newblock


\bibitem[\protect\citeauthoryear{Lai, Huang, Ahuja, and Yang}{Lai
  et~al\mbox{.}}{2017b}]%
        {lai2017deep}
\bibfield{author}{\bibinfo{person}{Wei-Sheng Lai}, \bibinfo{person}{Jia-Bin
  Huang}, \bibinfo{person}{Narendra Ahuja}, {and} \bibinfo{person}{Ming-Hsuan
  Yang}.} \bibinfo{year}{2017}\natexlab{b}.
\newblock \showarticletitle{Deep laplacian pyramid networks for fast and
  accurate super-resolution}. In \bibinfo{booktitle}{\emph{Proc. IEEE Conf.
  Comput. Vis. Pattern Recognit.}} \bibinfo{pages}{624--632}.
\newblock


\bibitem[\protect\citeauthoryear{Ledig, Theis, Huszar, Caballero, Cunningham,
  Acosta, Aitken, Tejani, Totz, Wang, and Shi}{Ledig et~al\mbox{.}}{2017}]%
        {Ledig_2017_CVPR}
\bibfield{author}{\bibinfo{person}{Christian Ledig}, \bibinfo{person}{Lucas
  Theis}, \bibinfo{person}{Ferenc Huszar}, \bibinfo{person}{Jose Caballero},
  \bibinfo{person}{Andrew Cunningham}, \bibinfo{person}{Alejandro Acosta},
  \bibinfo{person}{Andrew Aitken}, \bibinfo{person}{Alykhan Tejani},
  \bibinfo{person}{Johannes Totz}, \bibinfo{person}{Zehan Wang}, {and}
  \bibinfo{person}{Wenzhe Shi}.} \bibinfo{year}{2017}\natexlab{}.
\newblock \showarticletitle{Photo-Realistic Single Image Super-Resolution Using
  a Generative Adversarial Network}. In \bibinfo{booktitle}{\emph{The IEEE
  Conference on Computer Vision and Pattern Recognition (CVPR)}}.
\newblock


\bibitem[\protect\citeauthoryear{Lim, Son, Kim, Nah, and Lee}{Lim
  et~al\mbox{.}}{2017}]%
        {lim2017enhanced}
\bibfield{author}{\bibinfo{person}{Bee Lim}, \bibinfo{person}{Sanghyun Son},
  \bibinfo{person}{Heewon Kim}, \bibinfo{person}{Seungjun Nah}, {and}
  \bibinfo{person}{Kyoung~Mu Lee}.} \bibinfo{year}{2017}\natexlab{}.
\newblock \showarticletitle{Enhanced deep residual networks for single image
  super-resolution}. In \bibinfo{booktitle}{\emph{The IEEE Conference on
  Computer Vision and Pattern Recognition (CVPR) Workshops}},
  Vol.~\bibinfo{volume}{1}. \bibinfo{pages}{3}.
\newblock


\bibitem[\protect\citeauthoryear{Maninis, Pont-Tuset, Arbel{\'a}ez, and
  Van~Gool}{Maninis et~al\mbox{.}}{2018}]%
        {maninis2018convolutional}
\bibfield{author}{\bibinfo{person}{Kevis-Kokitsi Maninis},
  \bibinfo{person}{Jordi Pont-Tuset}, \bibinfo{person}{Pablo Arbel{\'a}ez},
  {and} \bibinfo{person}{Luc Van~Gool}.} \bibinfo{year}{2018}\natexlab{}.
\newblock \showarticletitle{Convolutional oriented boundaries: From image
  segmentation to high-level tasks}.
\newblock \bibinfo{journal}{\emph{IEEE transactions on pattern analysis and
  machine intelligence}} \bibinfo{volume}{40}, \bibinfo{number}{4}
  (\bibinfo{year}{2018}), \bibinfo{pages}{819--833}.
\newblock


\bibitem[\protect\citeauthoryear{Martin, Fowlkes, Tal, and Malik}{Martin
  et~al\mbox{.}}{2001}]%
        {martin2001database}
\bibfield{author}{\bibinfo{person}{David Martin}, \bibinfo{person}{Charless
  Fowlkes}, \bibinfo{person}{Doron Tal}, {and} \bibinfo{person}{Jitendra
  Malik}.} \bibinfo{year}{2001}\natexlab{}.
\newblock \showarticletitle{A database of human segmented natural images and
  its application to evaluating segmentation algorithms and measuring
  ecological statistics}. In \bibinfo{booktitle}{\emph{Computer Vision, 2001.
  ICCV 2001. Proceedings. Eighth IEEE International Conference on}},
  Vol.~\bibinfo{volume}{2}. IEEE, \bibinfo{pages}{416--423}.
\newblock


\bibitem[\protect\citeauthoryear{P{\u{a}}tr{\u{a}}ucean, Gurdjos, and
  Von~Gioi}{P{\u{a}}tr{\u{a}}ucean et~al\mbox{.}}{2012}]%
        {puatruaucean2012parameterless}
\bibfield{author}{\bibinfo{person}{Viorica P{\u{a}}tr{\u{a}}ucean},
  \bibinfo{person}{Pierre Gurdjos}, {and} \bibinfo{person}{Rafael~Grompone
  Von~Gioi}.} \bibinfo{year}{2012}\natexlab{}.
\newblock \showarticletitle{A parameterless line segment and elliptical arc
  detector with enhanced ellipse fitting}.
\newblock In \bibinfo{booktitle}{\emph{Computer Vision--ECCV 2012}}.
  \bibinfo{publisher}{Springer}, \bibinfo{pages}{572--585}.
\newblock


\bibitem[\protect\citeauthoryear{Segall, Molina, and Katsaggelos}{Segall
  et~al\mbox{.}}{2003}]%
        {SMK2003}
\bibfield{author}{\bibinfo{person}{C.A. Segall}, \bibinfo{person}{R. Molina},
  {and} \bibinfo{person}{A.K. Katsaggelos}.} \bibinfo{year}{2003}\natexlab{}.
\newblock \showarticletitle{High-resolution images from low-resolution
  compressed video}.
\newblock \bibinfo{journal}{\emph{IEEE Signal Processing Mag.}}
  \bibinfo{volume}{20}, \bibinfo{number}{3} (\bibinfo{year}{2003}),
  \bibinfo{pages}{37--48}.
\newblock


\bibitem[\protect\citeauthoryear{Sendik and Cohen-Or}{Sendik and
  Cohen-Or}{2017}]%
        {sendik2017deep}
\bibfield{author}{\bibinfo{person}{Omry Sendik} {and} \bibinfo{person}{Daniel
  Cohen-Or}.} \bibinfo{year}{2017}\natexlab{}.
\newblock \showarticletitle{Deep correlations for texture synthesis}.
\newblock \bibinfo{journal}{\emph{ACM Transactions on Graphics (TOG)}}
  \bibinfo{volume}{36}, \bibinfo{number}{5} (\bibinfo{year}{2017}),
  \bibinfo{pages}{161}.
\newblock


\bibitem[\protect\citeauthoryear{Sun and Hays}{Sun and Hays}{2017}]%
        {sun2017super}
\bibfield{author}{\bibinfo{person}{Libin Sun} {and} \bibinfo{person}{James
  Hays}.} \bibinfo{year}{2017}\natexlab{}.
\newblock \showarticletitle{Super-resolution Using Constrained Deep Texture
  Synthesis}.
\newblock \bibinfo{journal}{\emph{arXiv preprint arXiv:1701.07604}}
  (\bibinfo{year}{2017}).
\newblock


\bibitem[\protect\citeauthoryear{Tai, Yang, and Liu}{Tai
  et~al\mbox{.}}{2017a}]%
        {tai2017image}
\bibfield{author}{\bibinfo{person}{Ying Tai}, \bibinfo{person}{Jian Yang},
  {and} \bibinfo{person}{Xiaoming Liu}.} \bibinfo{year}{2017}\natexlab{a}.
\newblock \showarticletitle{Image super-resolution via deep recursive residual
  network}. In \bibinfo{booktitle}{\emph{The IEEE Conference on Computer Vision
  and Pattern Recognition (CVPR)}}, Vol.~\bibinfo{volume}{1}.
\newblock


\bibitem[\protect\citeauthoryear{Tai, Yang, Liu, and Xu}{Tai
  et~al\mbox{.}}{2017b}]%
        {tai2017memnet}
\bibfield{author}{\bibinfo{person}{Ying Tai}, \bibinfo{person}{Jian Yang},
  \bibinfo{person}{Xiaoming Liu}, {and} \bibinfo{person}{Chunyan Xu}.}
  \bibinfo{year}{2017}\natexlab{b}.
\newblock \showarticletitle{Memnet: A persistent memory network for image
  restoration}. In \bibinfo{booktitle}{\emph{Proceedings of the IEEE Conference
  on Computer Vision and Pattern Recognition}}. \bibinfo{pages}{4539--4547}.
\newblock


\bibitem[\protect\citeauthoryear{Tai, Liu, Brown, and Lin}{Tai
  et~al\mbox{.}}{2010}]%
        {tai2010super}
\bibfield{author}{\bibinfo{person}{Yu-Wing Tai}, \bibinfo{person}{Shuaicheng
  Liu}, \bibinfo{person}{Michael~S Brown}, {and} \bibinfo{person}{Stephen
  Lin}.} \bibinfo{year}{2010}\natexlab{}.
\newblock \showarticletitle{Super resolution using edge prior and single image
  detail synthesis}. In \bibinfo{booktitle}{\emph{2010 IEEE Computer Society
  Conference on Computer Vision and Pattern Recognition(CVPR)}},
  Vol.~\bibinfo{volume}{00}. \bibinfo{pages}{2400--2407}.
\newblock
\urldef\tempurl%
\url{https://doi.org/10.1109/CVPR.2010.5539933}
\showDOI{\tempurl}


\bibitem[\protect\citeauthoryear{Timofte, Agustsson, Van~Gool, Yang, Zhang,
  Lim, Son, Kim, Nah, Lee, et~al\mbox{.}}{Timofte et~al\mbox{.}}{2017}]%
        {Timofte_2017_CVPR_Workshops}
\bibfield{author}{\bibinfo{person}{Radu Timofte}, \bibinfo{person}{Eirikur
  Agustsson}, \bibinfo{person}{Luc Van~Gool}, \bibinfo{person}{Ming-Hsuan
  Yang}, \bibinfo{person}{Lei Zhang}, \bibinfo{person}{Bee Lim},
  \bibinfo{person}{Sanghyun Son}, \bibinfo{person}{Heewon Kim},
  \bibinfo{person}{Seungjun Nah}, \bibinfo{person}{Kyoung~Mu Lee},
  {et~al\mbox{.}}} \bibinfo{year}{2017}\natexlab{}.
\newblock \showarticletitle{Ntire 2017 challenge on single image
  super-resolution: Methods and results}. In \bibinfo{booktitle}{\emph{Computer
  Vision and Pattern Recognition Workshops (CVPRW), 2017 IEEE Conference on}}.
  IEEE, \bibinfo{pages}{1110--1121}.
\newblock


\bibitem[\protect\citeauthoryear{Timofte, De~Smet, and Van~Gool}{Timofte
  et~al\mbox{.}}{2013}]%
        {timofte2013anchored}
\bibfield{author}{\bibinfo{person}{Radu Timofte}, \bibinfo{person}{Vincent
  De~Smet}, {and} \bibinfo{person}{Luc Van~Gool}.}
  \bibinfo{year}{2013}\natexlab{}.
\newblock \showarticletitle{Anchored neighborhood regression for fast
  example-based super-resolution}. In \bibinfo{booktitle}{\emph{Proceedings of
  the IEEE International Conference on Computer Vision}}.
  \bibinfo{pages}{1920--1927}.
\newblock


\bibitem[\protect\citeauthoryear{Timofte, De~Smet, and Van~Gool}{Timofte
  et~al\mbox{.}}{2014}]%
        {timofte2014a+}
\bibfield{author}{\bibinfo{person}{Radu Timofte}, \bibinfo{person}{Vincent
  De~Smet}, {and} \bibinfo{person}{Luc Van~Gool}.}
  \bibinfo{year}{2014}\natexlab{}.
\newblock \showarticletitle{A+: Adjusted anchored neighborhood regression for
  fast super-resolution}. In \bibinfo{booktitle}{\emph{Asian Conference on
  Computer Vision}}. Springer, \bibinfo{pages}{111--126}.
\newblock


\bibitem[\protect\citeauthoryear{Wang, Bovik, Sheikh, and Simoncelli}{Wang
  et~al\mbox{.}}{2004}]%
        {wang2004image}
\bibfield{author}{\bibinfo{person}{Zhou Wang}, \bibinfo{person}{Alan~C Bovik},
  \bibinfo{person}{Hamid~R Sheikh}, {and} \bibinfo{person}{Eero~P Simoncelli}.}
  \bibinfo{year}{2004}\natexlab{}.
\newblock \showarticletitle{Image quality assessment: from error visibility to
  structural similarity}.
\newblock \bibinfo{journal}{\emph{IEEE transactions on image processing}}
  \bibinfo{volume}{13}, \bibinfo{number}{4} (\bibinfo{year}{2004}),
  \bibinfo{pages}{600--612}.
\newblock


\bibitem[\protect\citeauthoryear{Xie and Tu}{Xie and Tu}{2015}]%
        {xie2015holistically}
\bibfield{author}{\bibinfo{person}{Saining Xie} {and} \bibinfo{person}{Zhuowen
  Tu}.} \bibinfo{year}{2015}\natexlab{}.
\newblock \showarticletitle{Holistically-nested edge detection}. In
  \bibinfo{booktitle}{\emph{Proceedings of the IEEE international conference on
  computer vision}}. \bibinfo{pages}{1395--1403}.
\newblock


\bibitem[\protect\citeauthoryear{Yldrm and Gungor}{Yldrm and Gungor}{2012}]%
        {YG2012}
\bibfield{author}{\bibinfo{person}{D. Yldrm} {and} \bibinfo{person}{O.
  Gungor}.} \bibinfo{year}{2012}\natexlab{}.
\newblock \showarticletitle{A novel image fusion method using IKONOS satellite
  images}.
\newblock \bibinfo{journal}{\emph{Journal of Geodesy and Geoinformation}}
  \bibinfo{volume}{1}, \bibinfo{number}{1} (\bibinfo{year}{2012}),
  \bibinfo{pages}{27--34}.
\newblock


\bibitem[\protect\citeauthoryear{Zeyde, Elad, and Protter}{Zeyde
  et~al\mbox{.}}{2010}]%
        {zeyde2010single}
\bibfield{author}{\bibinfo{person}{Roman Zeyde}, \bibinfo{person}{Michael
  Elad}, {and} \bibinfo{person}{Matan Protter}.}
  \bibinfo{year}{2010}\natexlab{}.
\newblock \showarticletitle{On single image scale-up using
  sparse-representations}. In \bibinfo{booktitle}{\emph{International
  conference on curves and surfaces}}. Springer, \bibinfo{pages}{711--730}.
\newblock


\bibitem[\protect\citeauthoryear{Zhang, Tian, Kong, Zhong, and Fu}{Zhang
  et~al\mbox{.}}{2018}]%
        {zhang2018residual}
\bibfield{author}{\bibinfo{person}{Yulun Zhang}, \bibinfo{person}{Yapeng Tian},
  \bibinfo{person}{Yu Kong}, \bibinfo{person}{Bineng Zhong}, {and}
  \bibinfo{person}{Yun Fu}.} \bibinfo{year}{2018}\natexlab{}.
\newblock \showarticletitle{Residual Dense Network for Image Super-Resolution}.
  In \bibinfo{booktitle}{\emph{CVPR}}.
\newblock


\bibitem[\protect\citeauthoryear{Zhou, Chen, Liu, and Tang}{Zhou
  et~al\mbox{.}}{2011}]%
        {zhou2011}
\bibfield{author}{\bibinfo{person}{Qiang Zhou}, \bibinfo{person}{Shifeng Chen},
  \bibinfo{person}{Jianzhuang Liu}, {and} \bibinfo{person}{Xiaoou Tang}.}
  \bibinfo{year}{2011}\natexlab{}.
\newblock \showarticletitle{Edge-preserving Single Image Super-resolution}. In
  \bibinfo{booktitle}{\emph{Proceedings of the 19th ACM International
  Conference on Multimedia}} \emph{(\bibinfo{series}{MM '11})}.
  \bibinfo{publisher}{ACM}, \bibinfo{address}{New York, NY, USA},
  \bibinfo{pages}{1037--1040}.
\newblock
\showISBNx{978-1-4503-0616-4}
\urldef\tempurl%
\url{https://doi.org/10.1145/2072298.2071932}
\showDOI{\tempurl}


\end{thebibliography}
\end{document}